\documentclass[10pt,twocolumn,letterpaper]{article}

\usepackage{cvpr}              

\definecolor{cvprblue}{rgb}{0.21,0.49,0.74}
\usepackage[pagebackref,breaklinks,colorlinks,allcolors=cvprblue]{hyperref}
\usepackage{color}
\usepackage{multirow}
\def\paperID{4829} 
\def\confName{CVPR}
\def\confYear{2025}

\title{Linear Attention Modeling for Learned Image Compression}

\author{
    Donghui Feng$^{1}$\thanks{Equal contribution.}, Zhengxue Cheng$^{1}$\footnotemark[1], Shen Wang$^{1}$, Ronghua Wu$^{2}$, Hongwei Hu$^{2}$, Guo Lu$^{1}$, Li Song$^{1}$\thanks{Corresponding author. Email: song\_li@sjtu.edu.cn} \\
    $^1$ Shanghai Jiao Tong University \quad $^2$ Ant Group\\
}

\begin{document}
\maketitle
\begin{abstract}
%
Recent years, learned image compression has made tremendous progress to achieve impressive coding efficiency. Its coding gain mainly comes from non-linear neural network-based transform and learnable entropy modeling. However, most studies focus on a strong backbone, and few studies consider a low complexity design. In this paper, we propose LALIC, a linear attention modeling for learned image compression. Specially, we propose to use Bi-RWKV blocks, by utilizing the Spatial Mix and Channel Mix modules to achieve more compact feature extraction, and apply the Conv based Omni-Shift module to adapt to two-dimensional latent representation. Furthermore, we propose a RWKV-based Spatial-Channel ConTeXt model (RWKV-SCCTX), that leverages the Bi-RWKV to modeling the correlation between neighboring features effectively.
To our knowledge, our work is the first work to utilize efficient Bi-RWKV models with linear attention for learned image compression. Experimental results demonstrate that our method achieves competitive RD performances by outperforming VTM-9.1 by -15.26\%, -15.41\%, -17.63\% in BD-rate on Kodak, CLIC and Tecnick datasets. The code is available at \url{https://github.com/sjtu-medialab/RwkvCompress}.
\end{abstract}    
\section{Introduction}
\label{sec:intro}


Recently learned image compression (LIC) has achieved great success to realize efficient image transmission and storage by outperforming classical compression algorithms, including  JPEG~\cite{Wallace.1992.JPEG}, JPEG2000~\cite{Skodras.2001.JP2K}, High Efficiency Video Coding (HEVC/265)~\cite{Sullivan.2012.HEVC} and Versatile Video Coding (VVC/266)~\cite{Bross.2021.VVC}.
Mainstream LIC methods typically follow the non-linear transform coding framework \cite{Balle.2021.NTC}, combining learned entropy models with transform networks. Some pioneer LIC methods have investigated the adaptive context modeling, such as Hyerprior~\cite{Balle.2021.NTC}, spatialautoregressive model~\cite{Minnen.2018.Joint}, spatial checkerboard model, channel autoregressive models~\cite{Minnen.2020.Charm} or their alternative combinations~\cite{He.2022.ELIC}. Other representative LIC methods have explored better nonlinear analysis and synthesis transform networks, including widely-used residual convolution blocks~\cite{Cheng.2020.LIC}, swin-transformer~\cite{Liu.2021.SwinT}, and the mixture models of the above blocks~\cite{Liu.2023.TCM, Li.2023.FAT}, and even emerging Mamba blocks~\cite{Qin2024MambaVCLV}. 
The development of entropy models and transform networks have largely boosted the coding performance of learned image compression. 

\begin{figure}
  \includegraphics[width=1.0\linewidth]{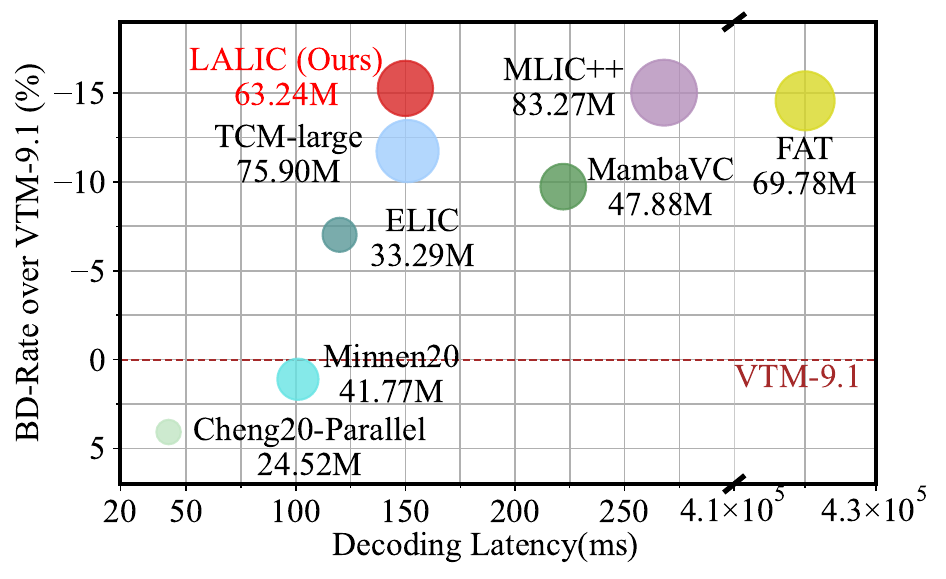}
  \caption{BD-rate vs Decoding Latency on Kodak dataset, where our proposed LALIC achieves the competitive BD-rate with moderate complexity. The Left-Top is better.} 
  \label{fig:teaser}
\end{figure}

%
%
Despite significant progress, each percent of coding gain in LIC has come with increased computational complexity. Meanwhile, transformers now have established as the mainstream backbone. To mitigate the quadratic complexity growth of transformers with sequence length, linear attention mechanisms were introduced to capture long-range dependencies with linear complexity, significantly reducing computational costs. Architectures such as Mamba\cite{Gu2023MambaLS} and RWKV\cite{Peng.2023.RWKV}, originally from natural language processing, have successfully expanded into computer vision as VMamba\cite{Liu2024VMambaVS} and Vision-RWKV\cite{Duan.2024.Vision-RWKV}. This linear complexity is particularly advantageous for low-level tasks that require pixel-wise processing of long sequences and global dependency capture in high-resolution images, achieving a global receptive field without complex strategies like window partitioning. Inspired by these advances, we propose utilizing linear attention in learned image compression.

In this work, we introduce LALIC, a novel Linear Attention-based Learned Image Compression architecture, leveraging RWKV's linear complexity advantage. Specifically, we propose Bi-RWKV blocks, utilizing Spatial-Mix and Channel-Mix modules to achieve more compact feature extraction, and incorporate Omni-Shift\cite{Yang.2024.Restore-RWKV} module to adapt to two-dimensional latent representations. Furthermore, we propose a RWKV-based Spatial-Channel ConTeXt Models (RWKV-SCCTX), that leverages the Bi-RWKVs to modeling the correlation between neighboring features effectively, to further improve the RD performance.
Experimental results demonstrate our proposed architecture achieves competitive rate-distortion performance with faster decoding speed in terms of widely-used PSNR and MS-SSIM quality metrics. 

To our knowledge, our work is the first work to utilize efficient RWKV models with linear attention for learned image compression. 
RWKVs are designed for fast inference and shows competitive compression performance compared to the state-of-the-art approaches as illustrated in Figure~\ref{fig:teaser}. In summary, our contributions are listed as follows.
\begin{itemize}
    \item We propose a new transform backbone with Bi-RWKV Blocks, which utilize the Spatial-Mix and Time-Mix to achieve more compact features extraction and apply the Omni-Shift module to adapt to two-dimensional latent representation.
    \item We propose a RWKV-based Spatial-Channel ConTeXt model (RWKV-SCCTX), that leverages Bi-RWKVs to modeling the correlation between neighboring features effectively, to further improve the RD performance.
    \item Experimental results demonstrate that our method achieves competitive RD performances by outperforming VTM-9.1 by -15.26\%, -15.41\%, -17.63\% in BD-rate on Kodak, CLIC and Tecnick datasets.
\end{itemize}

\section{Related Works}
\label{sec:related}

\subsection{Learned Image Compression}

In the past decade, learned image compression has demonstrated remarkable potential and made a significant success. We review the prevailing methods from two aspects, transform network and entropy modeling, where transform network can be further categorized into CNN-based and Transformer-based models.


\noindent\textbf{CNN-based Models} Some early works typically utilize the convolution neural networks with generalized divisive normalization (GDN) layers \cite{Balle17, Balle.2018.Hyperprior, Minnen.2018.Joint} in the analysis and synthesis transform to achieved good performance in image compression. Following that, attention mechanisms and residual blocks \cite{Cheng.2020.LIC, Zhou.2019.EOI} were integrated into the VAE architecture to enhance the capability of feature extraction. However, the limited receptive field constrained the further development of these models.




\noindent\textbf{Transformer-based Models} Transformers have demonstrated notable success in various computer vision tasks and Swin-Transformer~\cite{Liu.2021.SwinT} is proposed to restricts self-attention to local windows while enabling cross-window connections to enlarge the global attention. Studies~\cite{Zhu.2021.TBTC, Lu.2022.TIC, Zou.2022.DDW} show that nonlinear transforms based on Swin-Transformer improve compression efficiency over CNNs. Building on this, Liu et al.\cite{Liu.2023.TCM} introduce the TCM Block, which combines transformers and CNNs to enhance non-local and local information aggregation. Similarly, Li et al.\cite{Li.2023.FAT} propose Frequency-aware transformer blocks that adaptively ensemble diverse frequency components captured through various window shapes. However, introducing transformer still bring large computation complexity overhead to learned image compression.



\noindent\textbf{Entropy Modeling} 
Entropy modeling is essential in learned image compression to eliminate the redundancy of latent features. Most previous methods leverage joint autoregressive and hyperprior models \cite{Minnen.2018.Joint}, categorized into spatial \cite{Minnen.2018.Joint} (SA), channel-wise \cite{Minnen.2020.Charm} (CA), or combined approaches \cite{He.2022.ELIC}. Recent works incorporate transformers to capture long-range dependencies, enhancing entropy precision. For instance, Qian~\emph{et al.}~\cite{Qian.2022.EntroFormer} applies global self-attention for spatial dependencies, while Koyuncu~\emph{et al.}~\cite{Koyuncu.2022.CtxFormer} uses spatial-channel attention to boost R-D performance. However, high memory and computational demands limit real-world applicability, especially for high-resolution images. Liu~\emph{et al.}~\cite{Liu.2023.TCM} integrates Swin-Transformer within a channel-wise model for added spatial dependency, though with limited R-D gains. Li~\emph{et al.}\cite{Li.2023.FAT} propose T-CA, focusing on optimized channel-wise attention.

\subsection{Linear Attention Models}


Several architectures with linear attention have been designed for fast training and inference. Here we mainly introduce two representative models, Mamba and RWKV. 

\noindent\textbf{Mamba}
Mamba\cite{Gu2023MambaLS, Zhu2024VisionME}, a robust sequence model grounded in state-space models (SSMs), has emerged as a prominent contender to traditional Transformers. Additionally, the introduction of VMamba\cite{Liu2024VMambaVS}, leveraging SS2D, has further enhanced the capabilities of this model. Recent studies have underscored the superior performance of Mamba-based models over Transformer-based counterparts\cite{Guo2024MambaIRAS, Ji2024DeformMambaNF}, particularly in tasks such as image classification and multimodal learning\cite{Chen2024VideoMS, Qiao2024VLMambaES, Ma2024UMambaEL}. 
Mamba has also been introduced to image compression tasks. MambaVC\cite{Qin2024MambaVCLV} builds an efficient compression network based on SSM, capturing informative global contexts.







\begin{figure*}[t]
    \centering
    \begin{minipage}[b]{0.68\textwidth}
        \centering
        \subcaptionbox{Linear Attention based Learned Image Compression.\label{fig:overview}}{
            \raisebox{0.3cm}
            {\includegraphics[width=\textwidth]{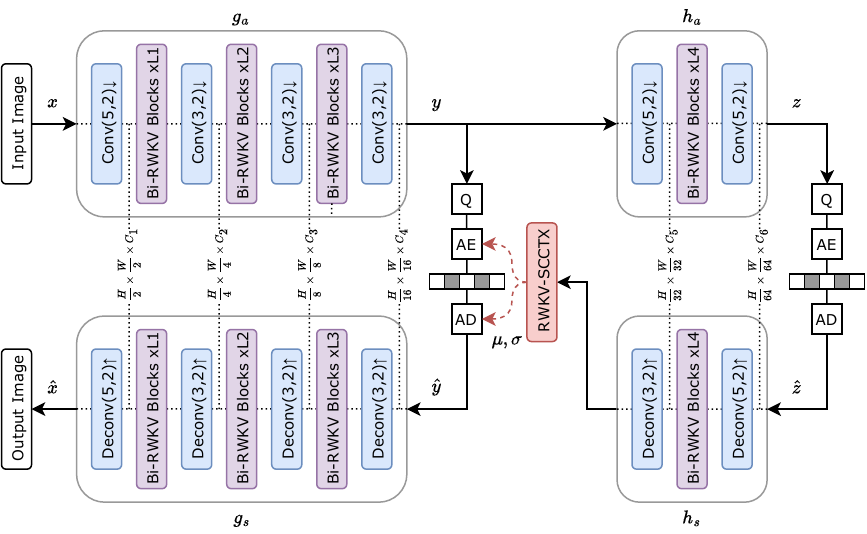}}}
    \end{minipage}%
    \hfill
    \begin{minipage}[b]{0.3\textwidth}
        \centering
        \subcaptionbox{Bi-RWKV Block.\label{fig:rwkv-block}}{
            \includegraphics[width=\textwidth]{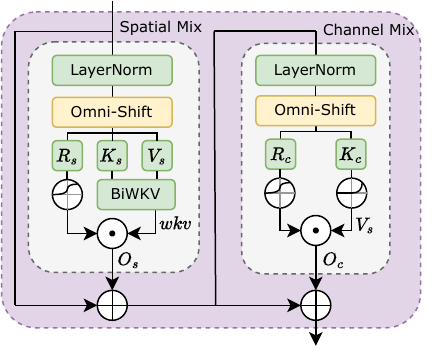}}
        \hfill
        \vspace{0.1cm}
        \subcaptionbox{Omni-Shift Layer.\label{fig:omni-shift}}{
            \includegraphics[width=\textwidth]{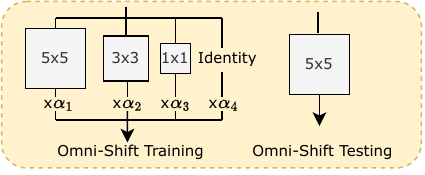}}
    \end{minipage}
    \caption{(a) Overview of proposed Linear Attention based Learned Image Compression (LALIC). Conv$(N,2)\downarrow$ and Deconv$(N,2)\uparrow$ represent strided down convolution and strided up convolution with $N \times N$ filters, respectively. There are $L$ identical Bi-RWKV Blocks stacked after downsample or upsample conv layer. AE, AD, and Q represent Arithmetic Encoding, Arithmetic Decoding, and Quantization. RWKV-SCCTX is the proposed RWKV-based Space-Channel Context model, illustrate in Fig.\ref{fig:entropy-model}. (b) The details of the Bi-RWKV Block. Omni-Shift\cite{Yang.2024.Restore-RWKV} denotes a reparameterized 5x5 depthwise convolution to capture local context. And BiWKV is the Bidirectional Attention proposed by \cite{Duan.2024.Vision-RWKV}.}
    \label{fig:main}
    \vspace{-15pt}
\end{figure*}

\noindent\textbf{RWKV}
The Receptance Weighted Key Value (RWKV) model~\cite{Bo.rwkv.2023, Datta2024TheEO} has emerged as an efficient alternative to Transformers.
RWKV offers distinctive advantages over standard Transformers, including a WKV attention mechanism for building long-range dependencies with linear computational complexity and a token shift layer to capture local context effectively. 
Recent study\cite{Peng2024EagleAF} shows that RWKV achieves performance on par with, or even exceeding that of both Transformers and Mamba in NLP tasks.
Vision-RWKV\cite{Duan2024VisionRWKVEA} has successfully transitioned RWKV's capabilities from NLP to vision tasks, demonstrating superior performance compared to vision Transformers by incorporating bidirectional WKV attention and quad-directional token shift mechanisms to harness spatial information in 2D images efficiently.
Several models derived from RWKV and Vision-RWKV have been developed and applied to diverse vision-related tasks. For instance, RWKV-SAM\cite{Yuan2024MambaOR} is designed for general image segmentation, while BSBP-RWKV\cite{Zhou2024BSBPRWKVBS} is specialized for medical image segmentation. Diffusion-RWKV\cite{Fei2024DiffusionRWKVSR} has been introduced for image generation, effectively reducing spatial aggregation complexity, while Restore-RWKV\cite{Yang2024RestoreRWKVEA} is dedicated to medical image restoration.

\section{Methods}

\subsection{Overview}

Figure~\ref{fig:overview} illustrates the architecture of the proposed Linear Attention based Learned Image Compression (LALIC) model. Given a raw image $x$, the analysis transform $g_a(\cdot)$ maps it to a latent representation $y$, and further obtain the hyper latent $z$ using the hyper encoder $h_a$.

\begin{equation}
\begin{aligned}
& \boldsymbol{y}=g_a\left(\boldsymbol{x} ; \boldsymbol{\theta}_{g_a}\right) \\
& \boldsymbol{z}=h_a\left(\boldsymbol{y} ; \boldsymbol{\theta}_{h_a}\right)
\end{aligned}
\end{equation}

The quantized hyper latent $\hat z = Q(z)$ is entropy coded for rate $R(\hat{\boldsymbol{z}})$ with a learned factorized prior.
At the decoder side, we first use a hyper decoder $h_s$ to obtain the initial mean and variance:

\begin{equation}
(\tilde{\boldsymbol{\mu}}, \tilde{\boldsymbol{\sigma}})=h_s\left(\hat{\boldsymbol{z}} ; \boldsymbol{\theta}_{h_s}\right)
\end{equation}


Then, quantization operator $Q(\cdot)$ discretizes $y$ to $\hat y$ and we assume that $y$ follows a conditional Gaussian distribution:  
$p_{\hat{\boldsymbol{y}} \mid \hat{\boldsymbol{z}}}(\hat{\boldsymbol{y}} \mid \hat{\boldsymbol{z}}) \sim \mathcal{N}\left(\boldsymbol{\mu}, \boldsymbol{\sigma}^2\right)$, 
whose distribution parameters are predicted by a space-channel context (SCCTX) entropy model, which will be discussed in Section 3.3.
The rate of latent representation $R(\hat{\boldsymbol{y}})$ is computed by $\mathbb{E}\left[-\log _2\left(p_{\hat{\boldsymbol{y}} \mid \hat{\boldsymbol{z}}}(\hat{\boldsymbol{y}} \mid \hat{\boldsymbol{z}})\right)\right]$. Next, the decoder  $g_s$ is used to reconstruct image from the quantized latent $\hat{\boldsymbol{y}}$ :




%
%

\begin{equation}
\hat{\boldsymbol{x}}=g_s\left(\hat{\boldsymbol{y}} ; \boldsymbol{\theta}_{g_s}\right)
\end{equation}

Here we propose to use Bi-RWKV block for $g_a$, $g_s$, $h_a$, $h_s$ to enhance the backbone. Finally, we optimize the following training objectives:
\begin{equation}
 L = \lambda\|\boldsymbol{x}-\hat{\boldsymbol{x}}\|^2+R(\hat{\boldsymbol{z}})+R(\hat{\boldsymbol{y}})
\end{equation}
where $\lambda$ is the Lagrangian multiplier to control the rate-distortion trade-off.

\subsection{Bi-RWKV Transform Block}

Following the transformer based compression methods, we integrate a Bi-RWKV block within the nonlinear transforms \( g_a \), \( g_s \), \( h_a \), and \( h_s \). Each Bi-RWKV block follows the upsampling or downsampling operations in these transforms, as shown in Figure~\ref{fig:rwkv-block}. The Bi-RWKV block consists of two branches: a spatial mix and a channel mix branch.

\noindent\textbf{Spatial Mix}. Designed to capture long-range spatial dependencies, the spatial mix module takes an input feature map \( f \in \mathbb{R}^{H \times W \times C} \), which is reshaped into a sequence \( X \in \mathbb{R}^{T \times C} \), where \( T = H \times W \).


The Spatial-Mix module begins with layer normalization for stable training, followed by an \textit{Omni-Shift} operation to capture 2D local context. As shown in Figure~\ref{fig:omni-shift}, the shift operation is implemented as a depth-wise 5x5 convolution with a reparameterization trick: it uses multiple kernels during training, which are merged into a single kernel for inference. The shifted representation \( X_s \) is then passed through three linear layers to produce the receptance \( R_s \), key \( K_s \), and value \( V_s \). The \textit{BiWKV} attention mechanism subsequently computes global attention \( wkv \), modulated by the sigmoid-gated receptance \( \sigma(R_s) \) and projected to form the final output \( O_s \).

\noindent\textbf{Channel Mix}. The channel mix module performs feature fusion across channels, enhancing cross-channel information flow.


Starting with layer normalization and Omni-Shift, the channel mix module computes \( R_c \) and \( K_c \) via linear projections. Then, \( V_c \) is obtained from \( K_c \) using a squared ReLU activation, implicitly creating an MLP structure. The final output \( O_c \) is formed by modulating \( V_c \) with the sigmoid-gated \( R_c \), allowing efficient channel fusion and improved model capacity.

\noindent\textbf{BiWKV Attention}. The WKV attention mechanism is central to the Spatial-Mix module, allowing for efficient extraction of distant dependencies with linear complexity. Originating from the Attention-Free Transformer (AFT) \cite{Zhai.2021.AFT}, this approach uses linear $KV$ attention rather than the quadratic $QK$ attention. The attention for the \( t \)-th token in AFT is given by

\begin{equation}
kv_t = \frac{\sum_{t^{\prime}=1}^T \exp(K_{t^{\prime}}) \odot V_{t^{\prime}}}{\sum_{t^{\prime}=1}^T \exp(K_{t^{\prime}})}
\end{equation}
where the values $V$ captures exponential weighted contributions from all tokens of the keys $K$. The final output is computed as \( \sigma_q(Q_t) \odot kv_t \), where \( \sigma_q(Q_t) \) applies a sigmoid gating to the query \( Q_t \).

Building on the KV attention, WKV attention~\cite{Peng.2023.RWKV} introduces channel-wise decay parameters \( w \) and \( u \). Beyond the contributions from \( K \), the parameter \( u \) amplifies the current token, while \( w \) decays the contributions of other tokens based on their distance. For visual tasks, the bidirectional WKV (BiWKV) attention~\cite{Duan.2024.Vision-RWKV} for the \( t \)-th token is defined as:

\begin{equation}
wkv_t = \frac{\sum_{i=1, i \neq t}^T e^{-(|t-i|-1) / T \cdot w + k_i} v_i + e^{u + k_t} v_t}{\sum_{i=1, i \neq t}^T e^{-(|t-i|-1) / T \cdot w + k_i} + e^{u + k_t}}
\end{equation}
where \( k_i \) and \( v_i \) represent the key and value for the \( i \)-th token. The introduced parameters allow the formulation to balance local and global dependencies effectively, adjusting token interactions based on their proximity.




\begin{figure}[t]
    \begin{subfigure}[b]{0.15\textwidth}
        \includegraphics[width=\textwidth]{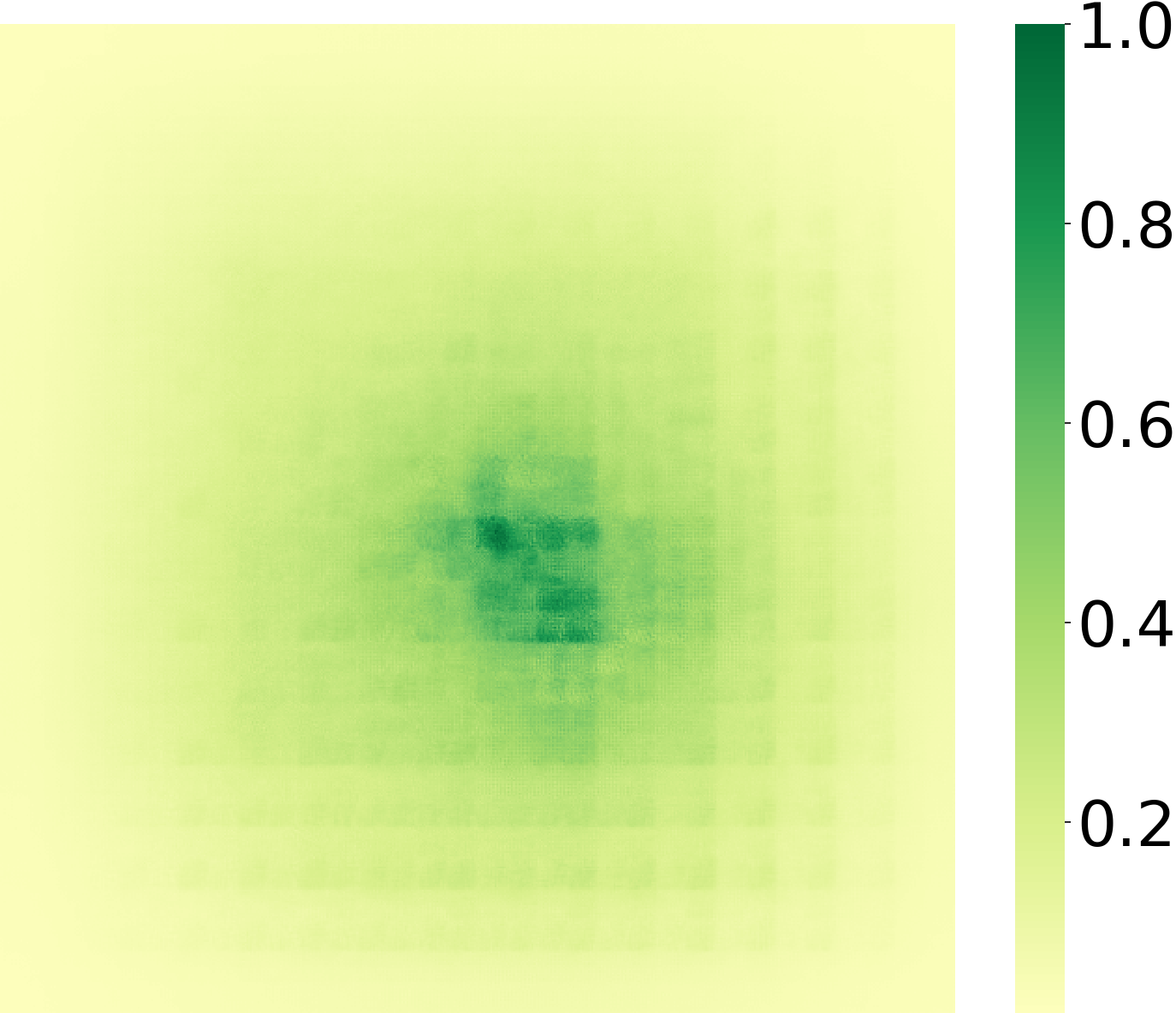}
        \caption{TCM}
        \label{fig:model-erf-sub1}
    \end{subfigure}
    \begin{subfigure}[b]{0.15\textwidth}
        \includegraphics[width=\textwidth]{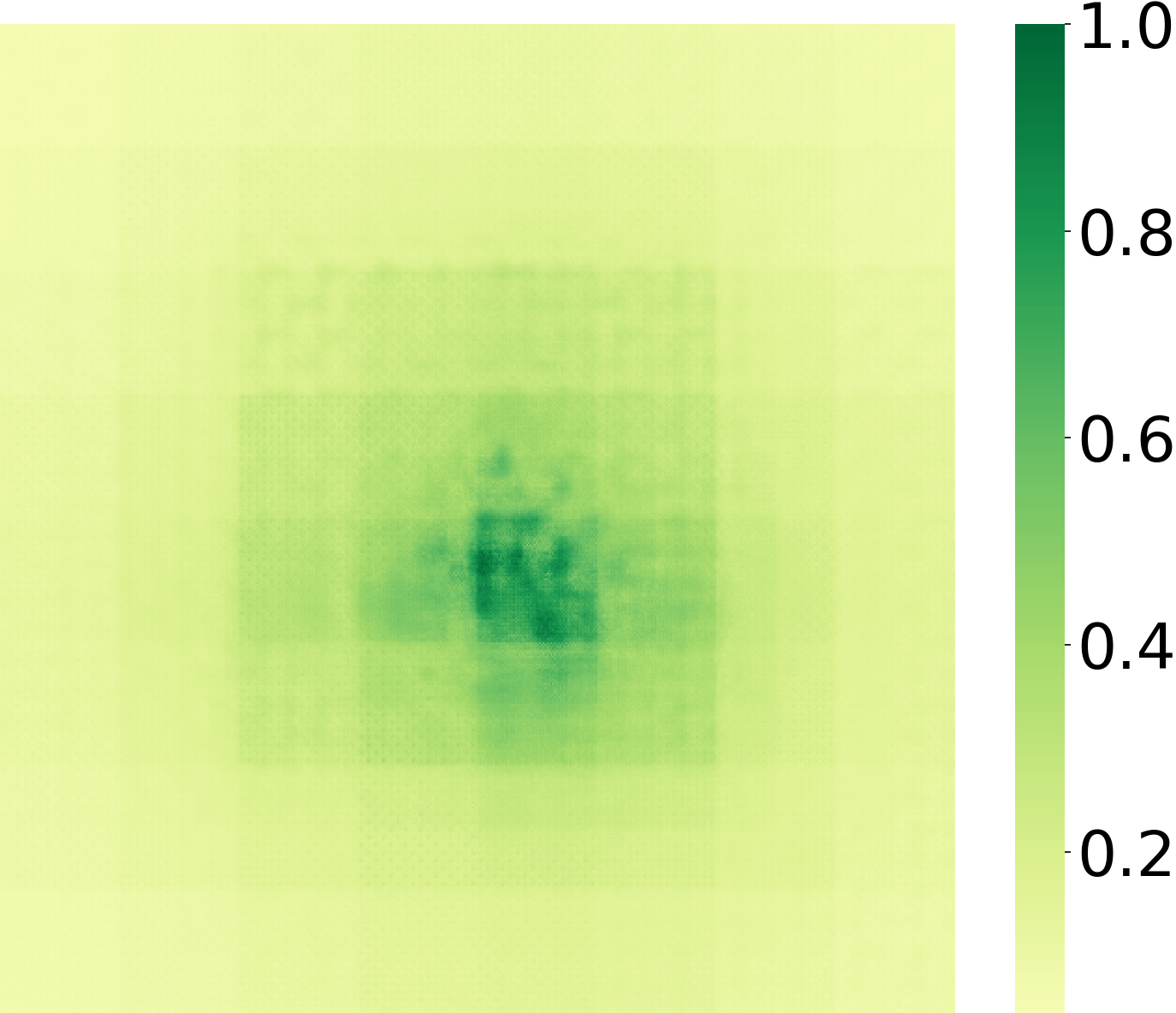}
        \caption{FAT}
        \label{fig:model-erf-sub2}
    \end{subfigure}
    \begin{subfigure}[b]{0.15\textwidth}
        \includegraphics[width=\textwidth]{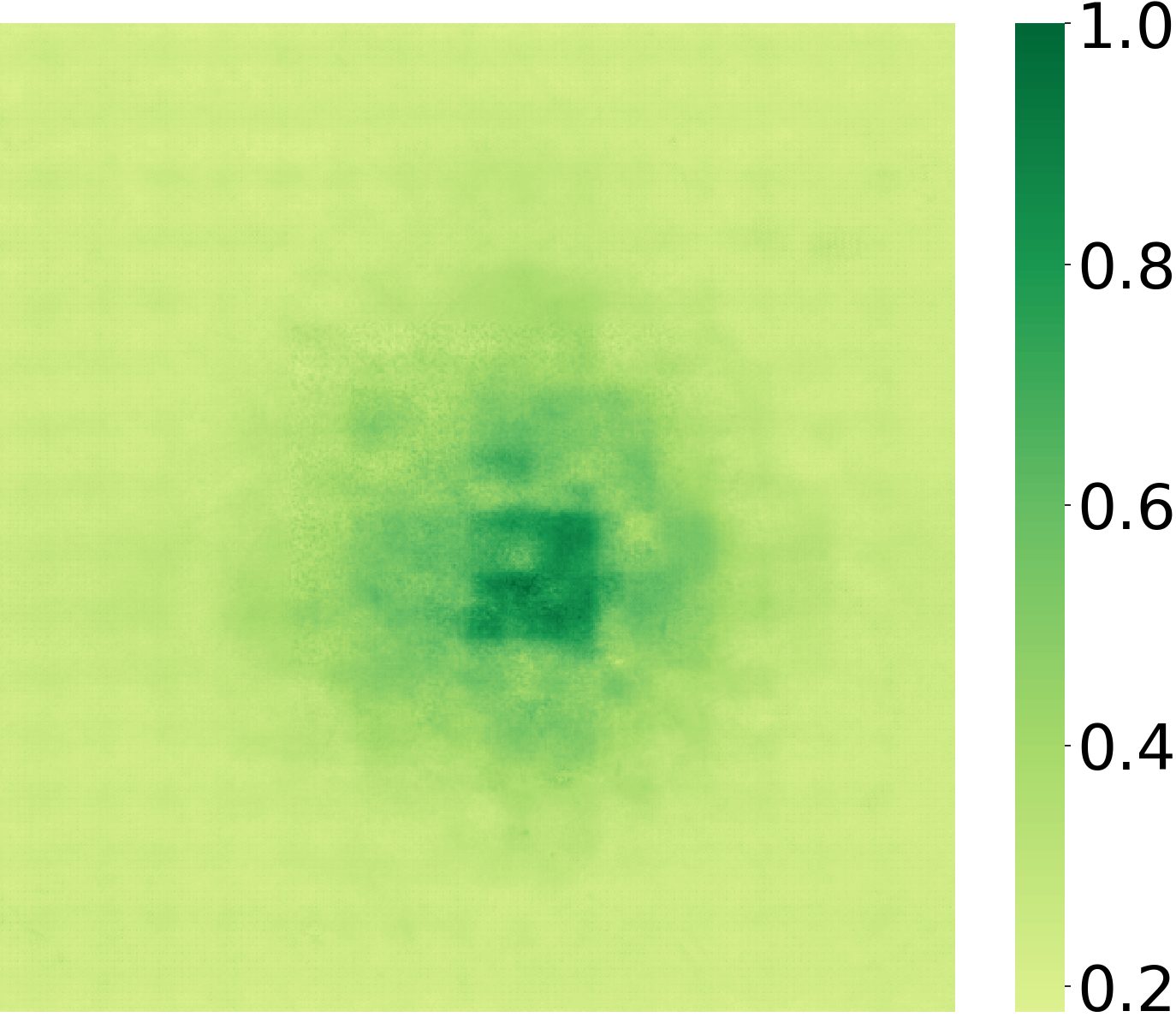}
        \caption{Bi-RWKV}
        \label{fig:model-erf-sub3}
    \end{subfigure}
  \caption{The effective receptive field (ERF) \cite{Luo.2016.ERF} visualization for the forward pass ($g_a$ \& $h_a$) of different models. A more extensively distributed dark area indicates a larger ERF.} 
  \label{fig:model-erf}
\end{figure}

\noindent\textbf{Effective Receptive Field.} The effective receptive field (ERF)~\cite{Luo.2016.ERF} describes how gradients flow from a latent location to the input image, defining the area it can perceive. A larger ERF enables the network to capture information from a broader area, which is particularly advantageous for nonlinear encoders in reducing redundancy. 

As shown in Figure~\ref{fig:model-erf}, we visualize the ERF of recent LIC models. The TCM block~\cite{Liu.2023.TCM} shows a shifted window pattern, while the FAT block~\cite{Li.2023.FAT} exhibits a locally enhanced window pattern due to its block-wise FFT design. In contrast, the RWKV block achieves a global ERF, enabling it to leverage a wider range of pixels for more effective redundancy elimination.

\subsection{RWKV Spatial-Channel Context Model}

\begin{figure}[t]
  \centering
  \includegraphics[width=0.95\linewidth]{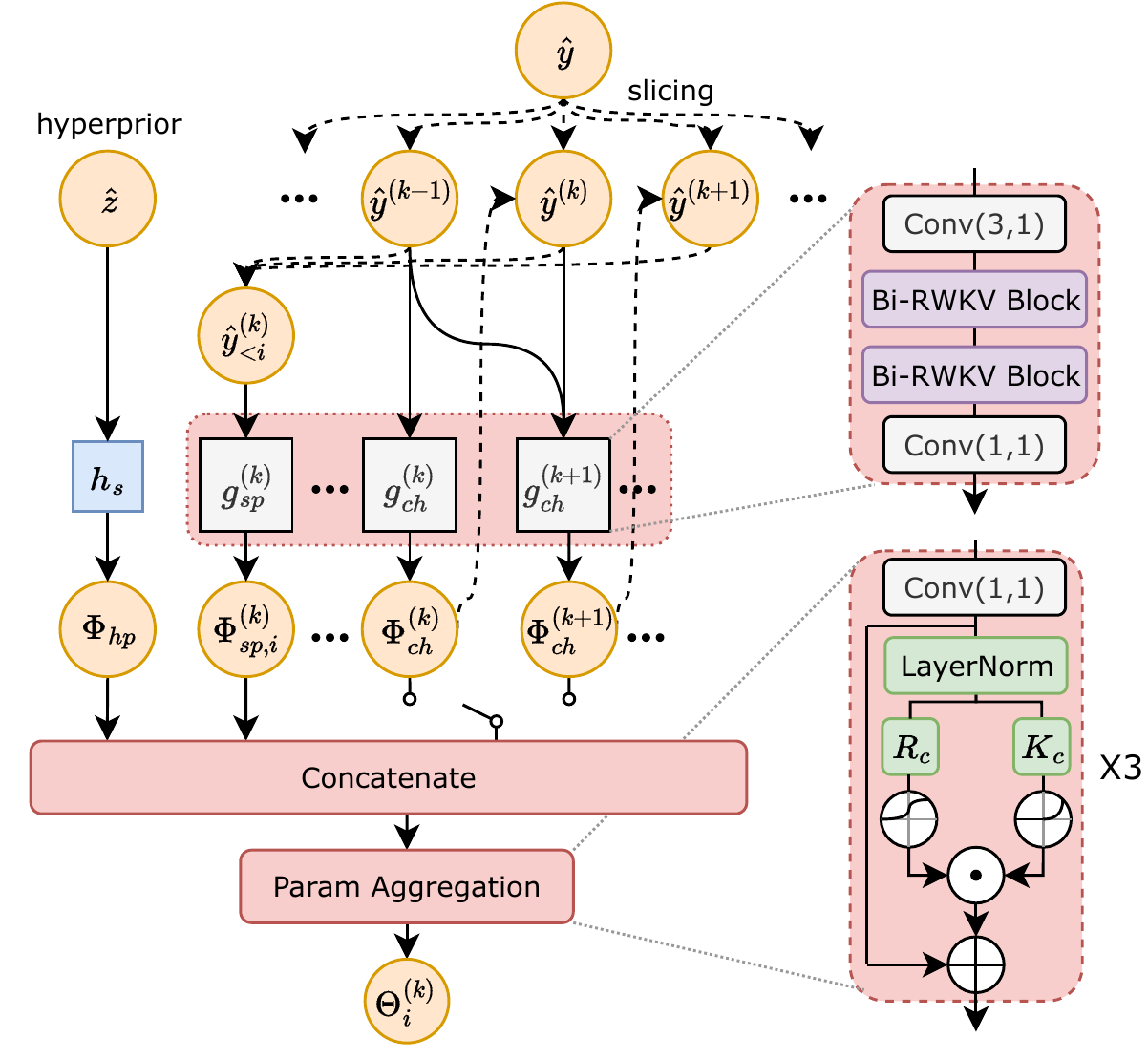}
  \caption{Diagram of the RWKV Spatial-Channel Context Model.} 
  \label{fig:entropy-model}
\end{figure}

The latent representation \( y \) in the transformed domain retains redundancies along both spatial and channel axes. Leveraging these redundancies, our entropy model correlates current decoding symbols with previously decoded ones to reduce the bit rate further. Motivated by recent advancements \cite{He.2022.ELIC, Liu.2023.TCM}, we propose an RWKV-based Spatial-Channel Context Model (RWKV-SCCTX) to model the conditional distribution of the latent variables more effectively, as shown in Fig.~\ref{fig:entropy-model}.

In the spatial dimension, we employ a checkerboard approach to divide symbols into two groups: anchors and non-anchors. The anchor group is encoded first, and the context derived from it is then used to encode the non-anchor group, capturing spatial dependencies.

For the channel dimension, we partition the channels into \( K \) chunks to build a channel-wise context:

\begin{equation}
   \Phi_{\mathrm{ch}}^{(k)} = g_{\mathrm{ch}}^{(k)}\left(\hat{\boldsymbol{y}}^{<k}\right), \quad k=2, \ldots, K 
\end{equation}
where \( \hat{\boldsymbol{y}}^{<k} = \left\{\hat{\boldsymbol{y}}^{(1)}, \ldots, \hat{\boldsymbol{y}}^{(k-1)}\right\} \) denotes the previously decoded channel chunks. Since the initial chunks are referenced more frequently by subsequent chunks, they tend to carry the majority of essential information. Allocating fewer channels to the initial chunks helps establish a more precise conditional distribution. Thus, for a latent representation \( \hat{\boldsymbol{y}} \) with \( M \) channels, we divide it into 5 chunks \( \hat{\boldsymbol{y}}^{(1)}, \ldots, \hat{\boldsymbol{y}}^{(5)} \) with channel allocations of \{16, 16, 32, 64, \( M - 128 \)\}, respectively. 

As illustrated in Fig.~\ref{fig:entropy-model}, for the \(k\)-th chunk’s \(i\)-th part, we apply a spatial context model \( g_{\mathrm{sp}}^{(k)} \) using checkerboard-masked convolutions to capture spatial context. If the part serves as an anchor, a zero context is used. Then, an RWKV-based network \( g_{\mathrm{ch}} \) is employed to model the channel-wise context \( \boldsymbol{\Phi}_{\mathrm{ch}}^{(k)} \) using decoded chunks. 

The spatial context and channel context are concatenated with the hyperprior context \( \boldsymbol{\Phi}_{hp} \). In the parameter aggregation network, this combined context is fused in a location-wise manner to predict the Gaussian distribution parameters, \( \boldsymbol{\Theta}_i^{(k)} = (\boldsymbol{\mu}_i^{(k)}, \boldsymbol{\sigma}_i^{(k)} ) \). We utilize the Channel Mix module without the Omni-Shift layer, to retain the $1 \times 1$ receptive field for causal decoding. Using the predicted entropy parameters, the latent \( {\boldsymbol{y}}_i^{(k)} \) is coded as follows:

\begin{equation}
    \hat{\boldsymbol{y}}_i^{(k)} = \operatorname{round}( {\boldsymbol{y}}_i^{(k)} - \boldsymbol{\mu}_i^{(k)}) + \boldsymbol{\mu}_i^{(k)}
\end{equation}

The decoded symbol \( \hat{\boldsymbol{y}}_i^{(k)} \) is subsequently used as context for computing \( \boldsymbol{\Phi}_{\mathrm{sp},(i+1)}^{(k)} \) or \( \boldsymbol{\Phi}_{\mathrm{ch}}^{(k+1)} \), iteratively progressing until the entire \( \hat{\boldsymbol{y}} \) is encoded.

\section{Experiments}
\subsection{Experimental Setup}

\noindent\textbf{Training Details} \; 
For training, we utilize the first 400,000 images of the OpenImages dataset~\cite{Kuznetsova.2020.OpenImages}, which provides high-resolution images suitable for learned compression tasks. The proposed LALIC models are trained with a batch size of 8, using the Adam optimizer. 
%
%
For MSE-optimized models, the Lagrange multipliers are set to \{0.0025, 0.0035, 0.0067, 0.0130, 0.0250, 0.0483\}, and for MS-SSIM optimized models, they are set to \{2.40, 4.58, 8.73, 16.64, 31.73, 60.50\}. The model is first trained for 40 epochs with a learning rate of \(1 \times 10^{-4}\). Then, the learning rate is decayed to \(1 \times 10^{-5}\) for an additional 4 epochs. Finally, we fine-tune the model using 512$\times$512 image crops for an additional 4 epochs. All experiments are conducted on an NVIDIA GeForce RTX 4090 GPU.


\begin{figure*}[h!]
  \centering
  \begin{subfigure}[b]{0.48\textwidth}
    \includegraphics[width=\textwidth]{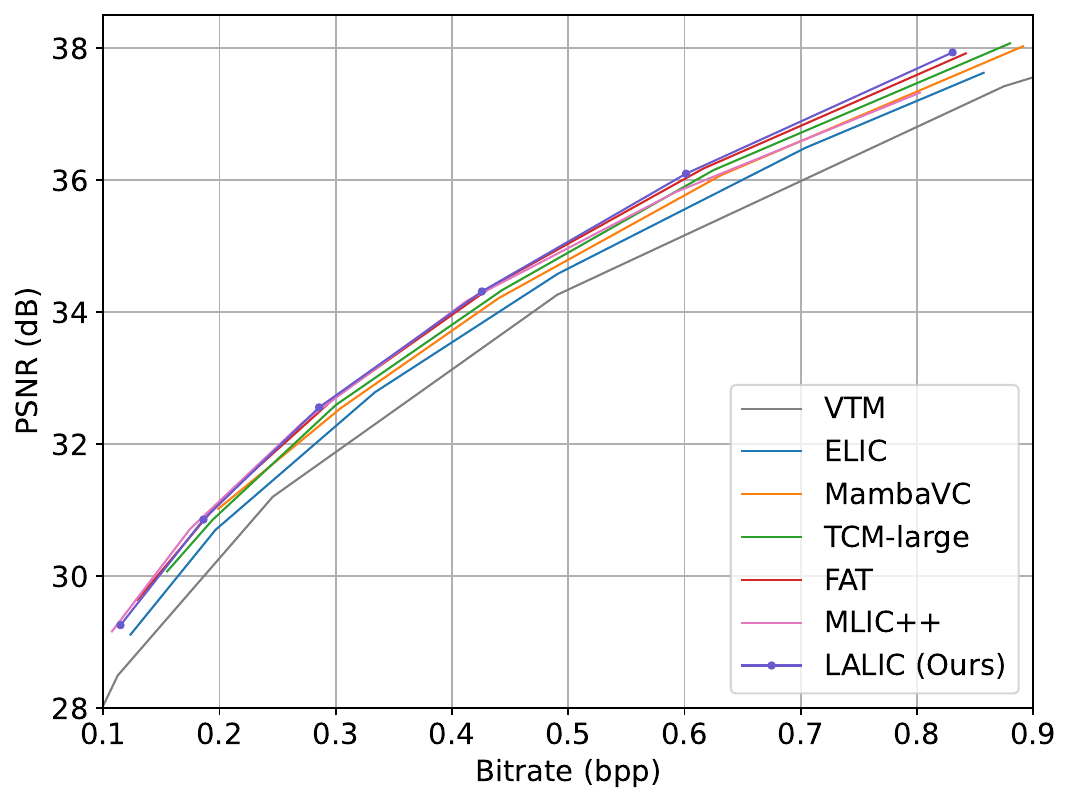}
    \label{fig:perf-kodak-psnr}
  \end{subfigure}
  \hfill
  \begin{subfigure}[b]{0.48\textwidth}
    \includegraphics[width=\textwidth]{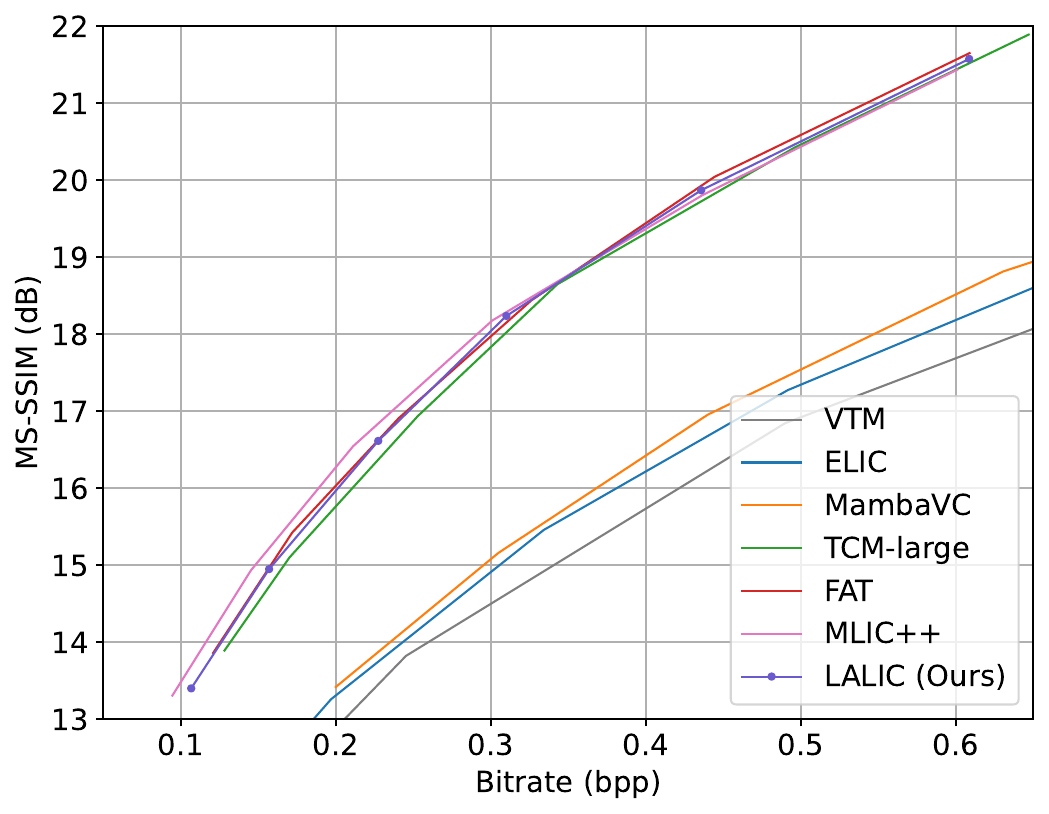}
    \label{fig:perf-kodak-msssim}
  \end{subfigure}
  \vspace{-1em}
  \caption{Rate-distortion performance on the Kodak dataset.}
  \label{fig:perf-kodak}
\end{figure*}

\begin{figure*}[htbp]
    \centering
    \begin{minipage}{0.48\linewidth}
        \centering
        \includegraphics[width=1\linewidth]{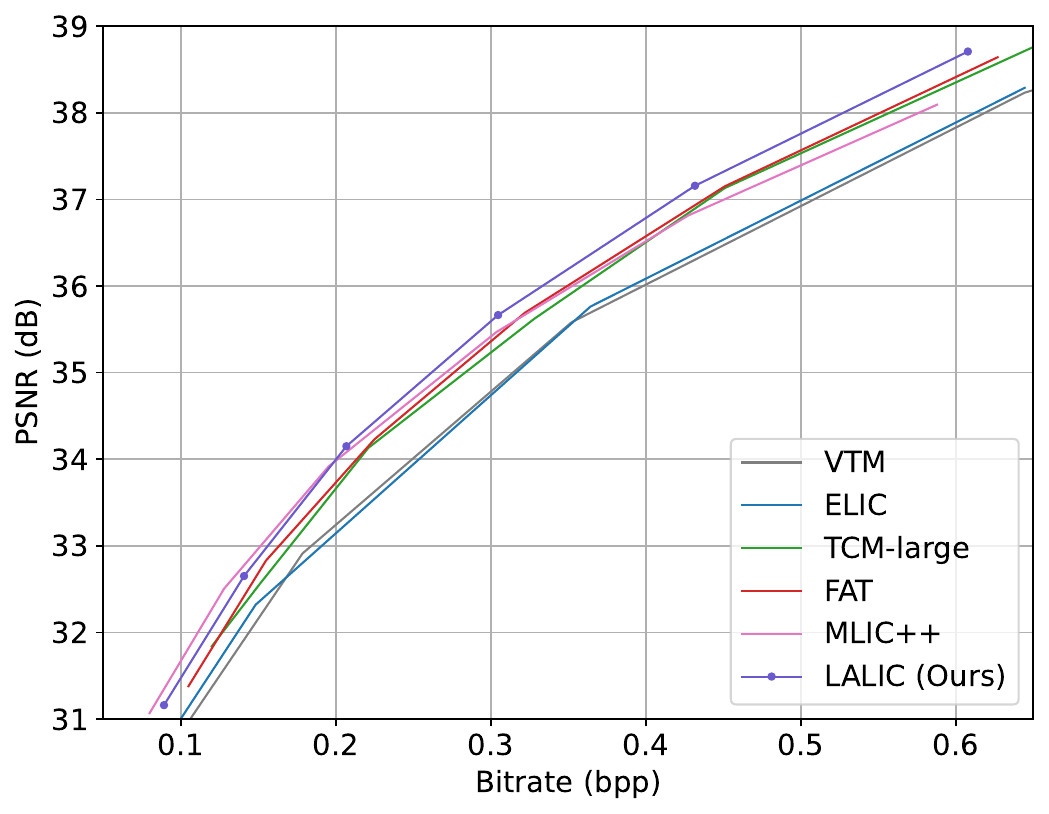}
        \caption{Rate-distortion performance on the CLIC dataset.}
        \label{fig:perf-clic}
    \end{minipage}
    \hfill
    \begin{minipage}{0.48\linewidth}
        \centering
        \includegraphics[width=1\linewidth]{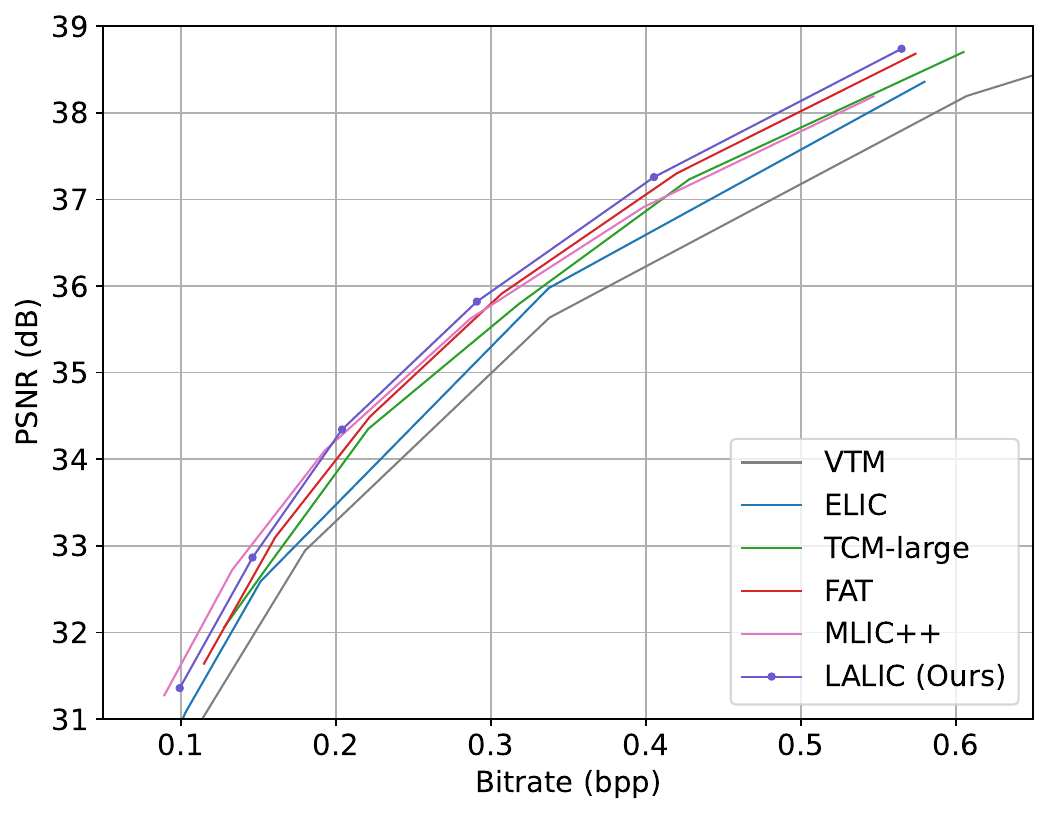}
        \caption{Rate-distortion performance on the Tecnick dataset.}
        \label{fig:perf-tecnick}
    \end{minipage}
\end{figure*}


\noindent\textbf{Model Settings} \;
For the RWKV blocks, we use a layer number $L_1$, $L_2$, $L_3$, $L_4$ with a configuration of \{2, 4, 6, 6\}. The latent feature \( y \) has a channel dimension \( M \) of 320, and the hyper-latent feature \( z \) has a channel dimension \( N \) of 192. To balance model capacity and computational complexity, the channels in the analysis and synthesis transforms (\(g_a\) and \(g_s\)) are set to $96$, $144$, and $256$, respectively. For the entropy model, we use $5$ slices, with $2$ RWKV blocks in each channel context. Additional hyper-parameters are discussed in the supplementary materials. 

\noindent\textbf{Evaluation Metrics} \;
We evaluate the proposed model on three datasets: (1) the Kodak dataset~\cite{Kodak.1993.Lossless}, containing 24 images with 768 × 512 resolution; (2) the Tecnick dataset~\cite{Asuni.2014.Tecnick}, containing 100 images at 1200 × 1200 resolution; and (3) the CLIC Professional Validation dataset~\cite{CLIC.2021.Workshop}, consisting of 41 images with resolutions up to 2K. Both Peak Signal-to-Noise Ratio (PSNR) and Multi Scale Structure Similarity (MS-SSIM) are used to measure the distortion, while bits per pixel (BPP) is used to measure the bitrate.

\subsection{Rate-Distortion Performance}

We compare the rate-distortion (R-D) performance of our method with state-of-the-art approaches, including traditional image codecs like BPG and VTM-9.1, as well as recent learned image compression (LIC) models \cite{Balle.2018.Hyperprior, Cheng.2020.LIC, He.2022.ELIC, Liu.2023.TCM, Li.2023.FAT, Jiang.2023.MLICpp}. We use corresponding model checkpoints if available or R-D curves from paper to get the R-D points. Figure~\ref{fig:perf-kodak} presents the R-D performance on the Kodak dataset using PSNR and MS-SSIM metrics. In addition to Kodak, we evaluate PSNR on the Tecnick and CLIC datasets, as shown in Figure~\ref{fig:perf-clic} and Figure~\ref{fig:perf-tecnick}.

\begin{table*}[t]
    \centering

\begin{tabular}{lrrrrrrrr}
\hline
\multirow{2}{*}{Method}                   & \multirow{2}{*}{Enc(s)} & \multirow{2}{*}{Dec(s)} & \multirow{2}{*}{Mem(G)} & \multirow{2}{*}{FLOPs(G)} & \multirow{2}{*}{Params(M)} & \multicolumn{3}{c}{BD-Rate (PSNR)}                                                 \\ \cline{7-9} 
                                          &                         &                         &                         &                           &                            & \multicolumn{1}{c}{Kodak} & \multicolumn{1}{c}{CLIC} & \multicolumn{1}{c}{Tecnick} \\ \hline
VTM-9.1                                   & 171.633                 & 0.177                   & -                       & -                         & -                          & 0.00\%                    & 0.00\%                   & 0.00\%                      \\
Minnen20~\cite{Minnen.2020.Charm}   & 0.077                   & 0.101                   & 0.446                   & 208.95                    & 41.77                      & 1.11\%                    & -                        & -                           \\
Cheng20-Parallel~\cite{He.2022.ELIC} & 0.090                   & 0.042                   & 0.413                   & 369.70                    & 24.52                      & 4.08\%                    & -                        & -                           \\
ELIC~\cite{He.2022.ELIC}                  & 0.237                   & 0.120                   & 0.373                   & 332.42                    & 33.29                      & -7.02\%                   & -1.19\%                  & -7.64\%                     \\
MambaVC~\cite{Qin2024MambaVCLV}           & 0.229                   & 0.222                   & 5.813                   & 393.37                    & 47.88                      & -9.73\%                   & -                        & -                           \\ \hline
TCM-large~\cite{Liu.2023.TCM}             & 0.157                   & 0.151                   & 1.698                   & 700.66                    & 75.90                      & -11.73\%                  & -9.41\%                  & -10.93\%                    \\
FAT~\cite{Li.2023.FAT}                    & \textbf{0.140}          & \textgreater 10.000     & 1.076                   & \textbf{245.46}           & 69.78                      & -14.56\%                  & -10.79\%                 & -14.40\%                    \\
MLIC++~\cite{Jiang.2023.MLICpp}           & 0.190                   & 0.268                   & \textbf{0.630}          & 443.17                    & 83.27                      & -15.02\%                  & -14.45\%                 & -17.21\%                    \\ \hline
LALIC (Ours)                              & 0.274                   & \textbf{0.150}          & 0.841                   & 286.16                    & \textbf{63.24}             & \textbf{-15.26\%}         & \textbf{-15.41\%}        & \textbf{-17.63\%}           \\ \hline
\end{tabular}

    \caption{Rate-distortion (R-D) performance and computational complexity of various learned image compression models on the Kodak dataset, evaluated on an NVIDIA RTX 4090 GPU. Lower BD-Rate values indicate better R-D performance. Bold font denote the best performance among the recent SOTA methods. ”-” indicates an unavailable result. Note that we re-run the FAT method using the official github code. More details will be discussed in the supplementary materials.}
    \label{tab:performance}
\end{table*}


As summarized in Table~\ref{tab:performance}, we further evaluate the BD-rate (PSNR) of different LIC methods. Our approach achieves state-of-the-art (SOTA) performance, surpassing VTM-9.1 by 15.26\% in BD-rate on the Kodak dataset. Furthermore, it demonstrates superior performance on high-resolution datasets, including the CLIC dataset (2K resolution) and the Tecnick dataset (1K resolution). These results highlight the global modeling capability of RWKV, which excels in handling high-resolution images by effectively capturing long-range dependencies. Additional quantitative results and subjective visual comparisons are provided in the supplementary materials.

\subsection{Computational Complexity}

We evaluate the computational complexity of the proposed model across four metrics: encoding time, decoding time, inference GPU memory consumption, and forward FLOPs. These measurements provide a well-rounded assessment of complexity from various perspectives. As shown in Table~\ref{tab:performance}, our model demonstrates a competitive balance between efficiency and rate-distortion (R-D) performance.

\begin{table}[tb]
\centering
\begin{tabular}{@{}llrrr@{}}
\toprule
\#Layers  & SCCTX     & FLOPs/G & Params/M & BD-rate \\ \midrule
2,2,2,2   & Conv      & 163.95    & 27.61      & 0.00\%     \\
2,4,6,2   & Conv      & 233.87    & 35.64      & -2.50\%    \\
2,4,6,6   & Conv      & 239.20    & 42.57      & -1.68\%    \\
2,4,6,6   & Conv Plus & 304.23    & 62.08      & -2.74\%    \\
2,4,6,6   & RWKV      & 286.16    & 63.24      & \textbf{-3.50\%}    \\ \bottomrule
\end{tabular}
\caption{Ablation study on the effect of varying RWKV block numbers, and with Conv based or RWKV based entropy models, showing BD-rate gain on the Kodak dataset.}
\label{tab:ablation}
\end{table}

Our proposed LALIC model achieves competitive encoding and decoding times with the lowest parameters among recent methods that attain more than 10\% bitrate savings. Specifically, our model has an encoding time of 274 ms and a decoding time of 150 ms, reflecting a practical level of latency suitable for real-world applications. Furthermore, with 286.16 GFLOPs, our model demonstrates efficient computational complexity, validating the Bi-RWKV module’s strong and effective modeling capacity. This efficiency is particularly valuable in scenarios requiring both high compression performance and low computational overhead.

\subsection{Ablation Studies and Analysis}


Learned visual compression involves two critical steps for redundancy removal: the powerful nonlinear transform and the delicate conditionally factorized Gaussian prior distribution to decorrelate the latent representation \( y \). To evaluate the effectiveness of the proposed LALIC architecture, we conducted ablation studies to analyze the impact of the Bi-RWKV nonlinear transform modules and the RWKV-SCCTX entropy model. Additionally, visualizations were provided to offer deeper insights into the design choices. 



\noindent\textbf{Number of Bi-RWKV Blocks}. We first examine the effect of Bi-RWKV block count on model performance, aiming to understand the trade-off between model complexity and compression efficiency. As shown in Table~\ref{tab:ablation}, we use inference loss as a proxy for evaluating compression performance. Results indicate that increasing the number of RWKV blocks consistently improves performance, particularly when additional blocks are added to high-resolution stages. This suggests that a deeper configuration in early stages allows the model to capture more detailed spatial information, enhancing overall compression effectiveness.

\noindent\textbf{Entropy Model Configuration}. Building on a Conv based SCCTX entropy model \cite{He.2022.ELIC}, we further evaluate the effectiveness of our RWKV-SCCTX by introducing the Bi-RWKV block to model channel context and incorporating channel mix layers to modulate entropy parameters. Previous research \cite{Sheng.2024.NVC1B} underscores the importance of increasing context parameters to enhance compression performance. Following this insight, we define the Conv Plus SCCTX by expanding the dimension and depth of the context model. 

As shown in Table~\ref{tab:ablation}, the RWKV SCCTX achieves superior performance with nearly the same number of parameters as Conv Plus, while requiring fewer FLOPs. Compared to the baseline Conv SCCTX, the RWKV SCCTX delivers a significant reduction in BD-rate, demonstrating its capability to improve compression efficiency without excessively increasing computational complexity.

\begin{figure*}[h!]
  \centering
    \begin{subfigure}[b]{\textwidth}
        \includegraphics[width=\textwidth]{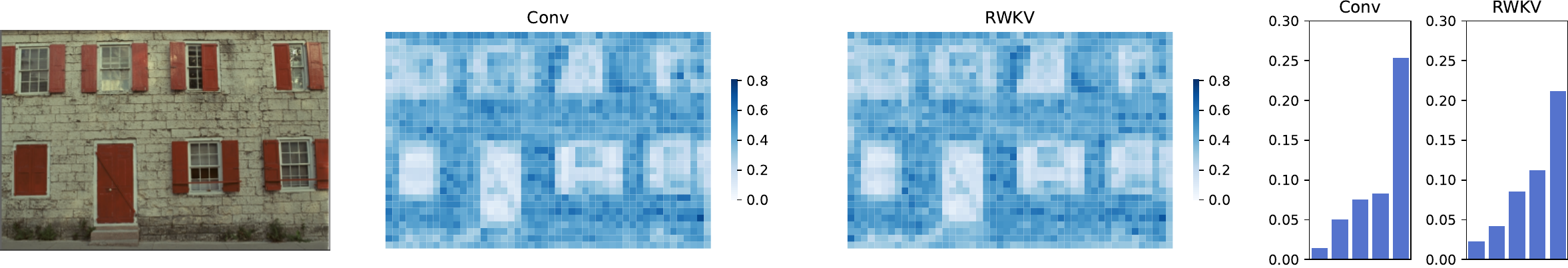}
        \label{fig:latent-sub1}
    \end{subfigure}
    \vspace{-6em}
    \begin{subfigure}[b]{\textwidth}
        \includegraphics[width=\textwidth]{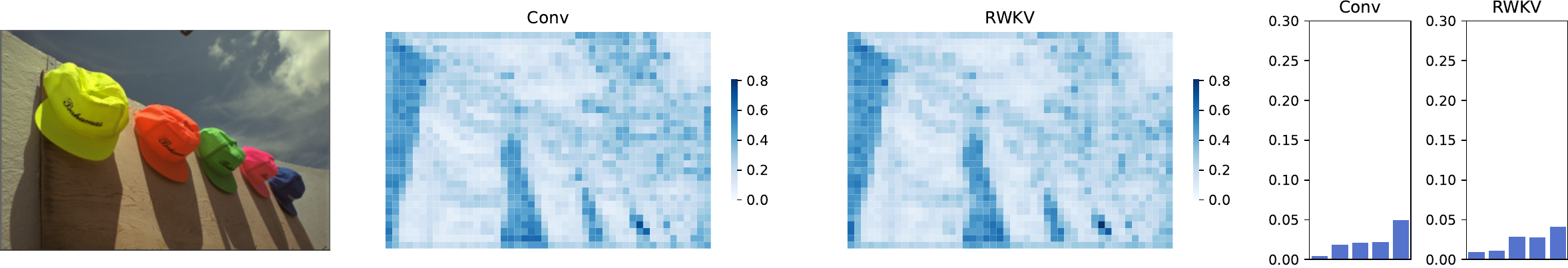}
        \label{fig:latent-sub2}
    \end{subfigure}
    \vspace{4em}
  \caption{Effectiveness of the RWKV-based entropy model in improving compression efficiency. The middle two columns present the scaled deviation maps from the SCCTX model using either Conv or RWKV. The right two columns illustrate the uneven entropy distribution across the 5 channel slices.}
  \label{fig:latent-scale}
\end{figure*}

\begin{figure}[h!]
    \centering
    \begin{subfigure}[b]{0.15\textwidth}
        \includegraphics[width=\textwidth]{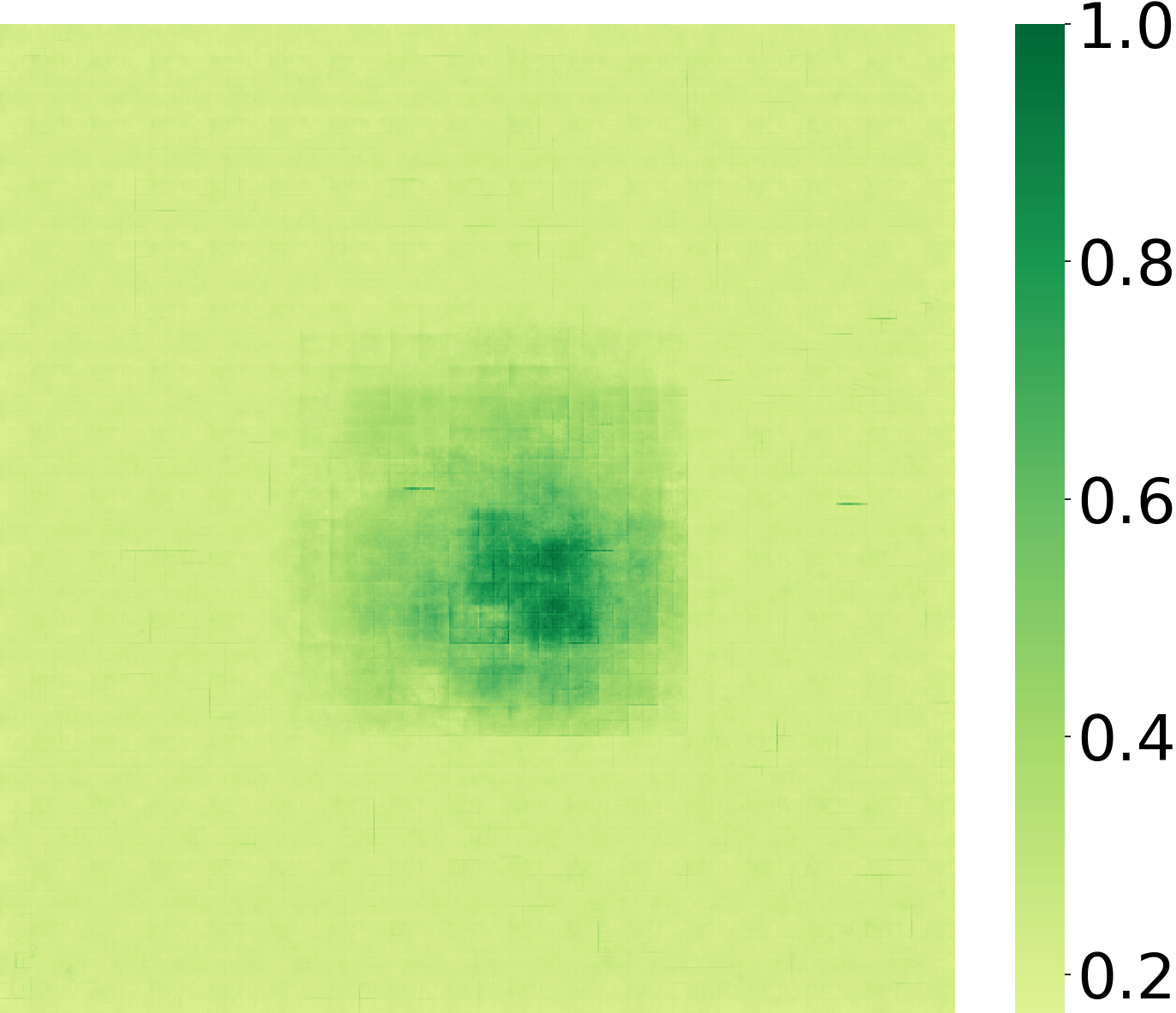}
        \caption{AFT}
        \label{fig:erf-sub1}
    \end{subfigure}
    \begin{subfigure}[b]{0.15\textwidth}
        \includegraphics[width=\textwidth]{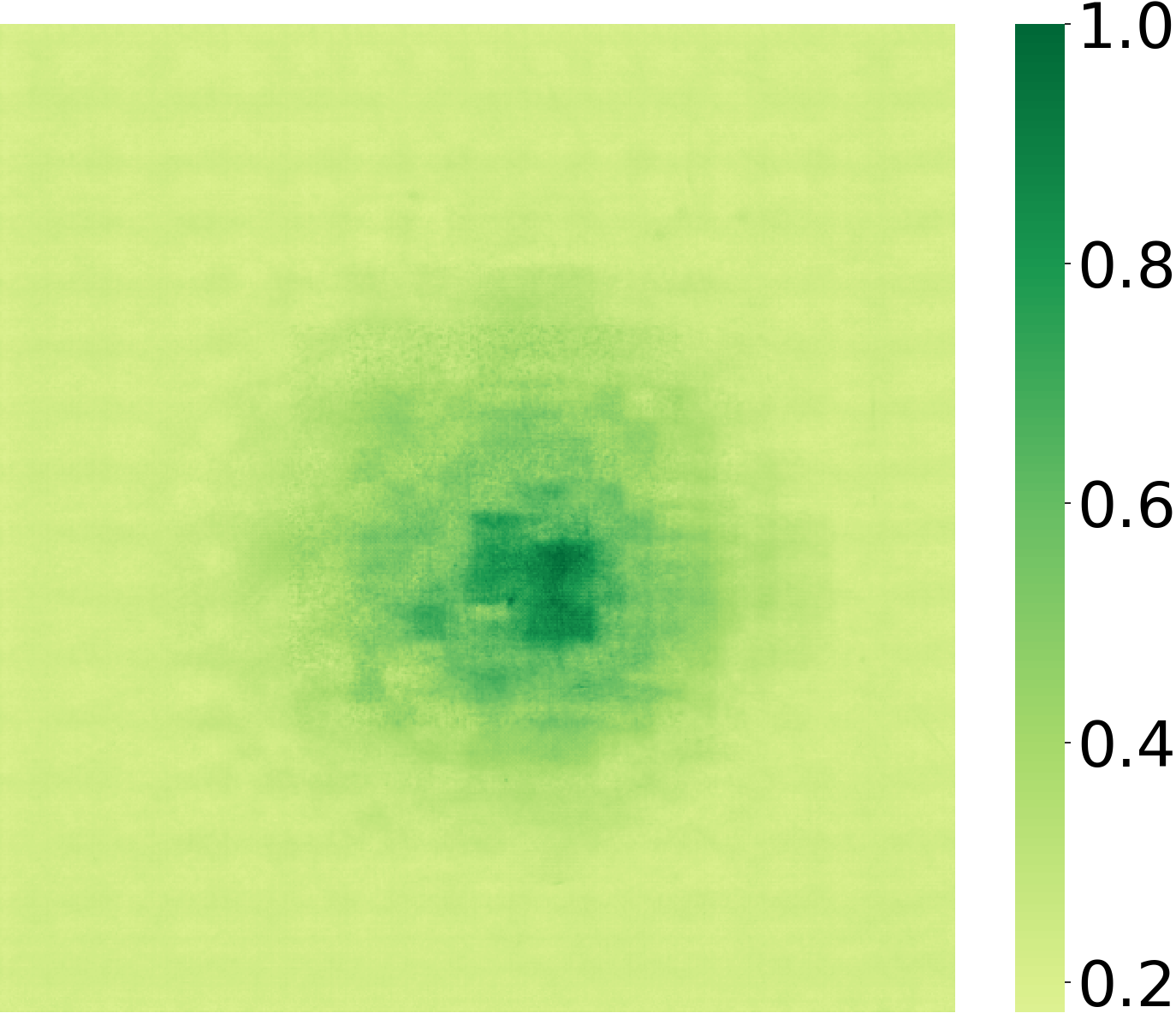}
        \caption{AFT+Shift}
        \label{fig:erf-sub2}
    \end{subfigure}
    \begin{subfigure}[b]{0.15\textwidth}
        \includegraphics[width=\textwidth]{fig/ERF-BiWKV-Shift.png}
        \caption{BiWKV+Shift}
        \label{fig:erf-sub3}
    \end{subfigure}
    \begin{subfigure}[b]{0.15\textwidth}
        \includegraphics[width=\textwidth]{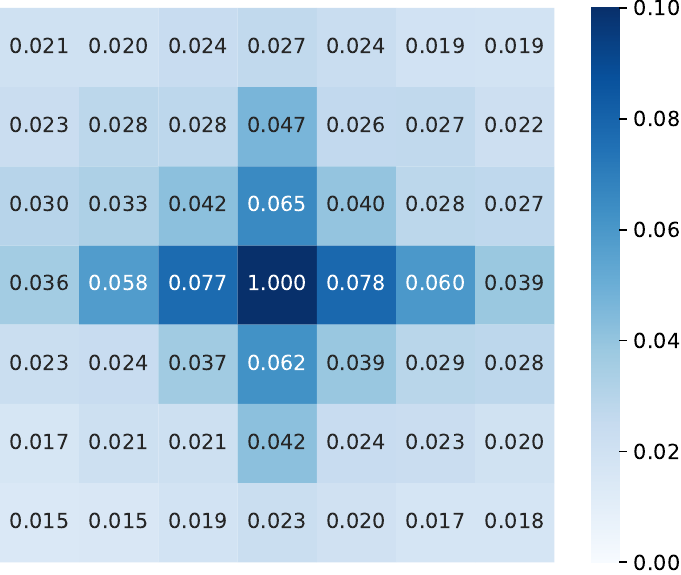}
        \caption{AFT}
        \label{fig:corr-sub1}
    \end{subfigure}
    \begin{subfigure}[b]{0.15\textwidth}
        \includegraphics[width=\textwidth]{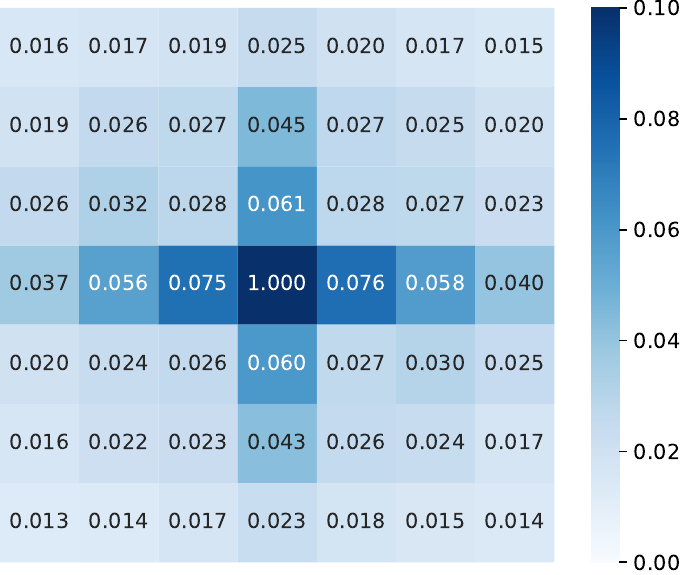}
        \caption{AFT+Shift}
        \label{fig:corr-sub2}
    \end{subfigure}
    \begin{subfigure}[b]{0.15\textwidth}
        \includegraphics[width=\textwidth]{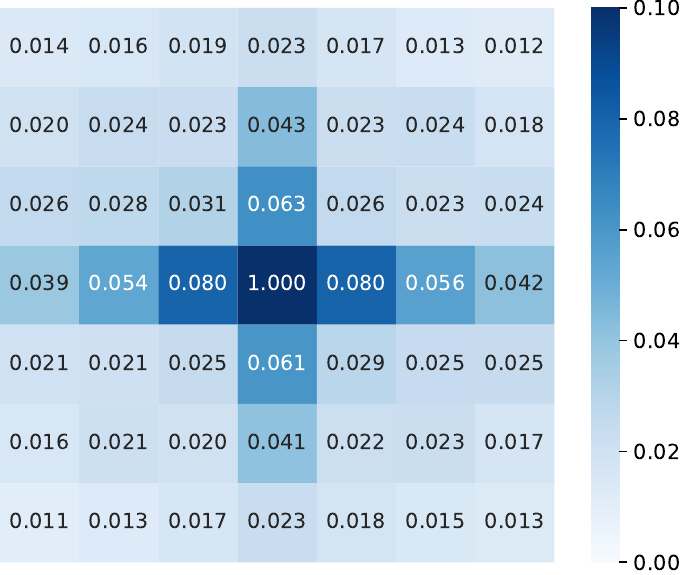}
        \caption{BiWKV+Shift}
        \label{fig:corr-sub3}
    \end{subfigure}
  \caption{Visualization of different attention mechanisms. The top row shows the effective receptive field (ERF) \cite{Luo.2016.ERF} for the forward pass ($g_a$ and $h_a$) of the constructed models. A more extensively distributed dark area indicates a larger ERF. The bottom row presents the local correlation matrix of the normalized latent representation \((y - \mu)/\sigma\). Each value represents the Pearson correlation coefficient between the vector at a given location and the center location, computed along the channel dimension and averaged across all images in the Kodak dataset.} 
  \label{fig:attn-erf}
\end{figure}

To gain further insights into the improvements offered by RWKV-based context modeling, we analyze the quantization loss in the latent domain and the entropy distribution of the latent representation. The information loss during compression is quantified by the scaled deviation~\cite{Xie.2021.EInv} between \( \hat{\boldsymbol{y}} \) and \( \boldsymbol{y} \), defined as \( \epsilon = \operatorname{abs}(\hat{\boldsymbol{y}} - \boldsymbol{y}) / \Sigma \boldsymbol{y} \).

Figure~\ref{fig:latent-scale} illustrates the scaled deviation map and channel entropy distributions from the Kodak dataset. The results clearly indicate that the RWKV block significantly reduces latent deviation and concentrates entropy more effectively in the initial slices. This behavior highlights the RWKV's ability to balance global and local dependencies, enabling superior entropy modeling and more efficient compression.

\noindent\textbf{Attention Mechanisms in the Block}. We investigate the contributions of various internal components within the RWKV block, focusing on different attention mechanisms and the inclusion of an Omni-Shift layer. Table~\ref{tab:block-component} highlights the impact of these design choices on computational complexity and compression performance. As the attention mechanisms evolve, performance improves with increasing FLOPs, while the growth in parameter count remains minimal. Refer to supplementary materials for comparisons with other linear attention mechanisms.


To demonstrate the influence of attention mechanisms on the network, Figure~\ref{fig:attn-erf} illustrates the effective receptive field (ERF) across different block configurations. These ERF visualizations reveal how distinct attention mechanisms and shift layers shape the receptive field, thereby impacting the model’s ability to capture contextual information. The shift layer effectively captures local context and broadens the receptive field, while transitioning from AFT to BiWKV enables the model to capture more global information. Furthermore, the enhanced attention mechanisms help reduce local correlations, as demonstrated by the Pearson correlation matrix.

\begin{table}[h!]
\centering
\begin{tabular}{@{}lrrr@{}}
\toprule
Attention              & $\Delta$FLOPs (G) & Params (M) & Loss   \\ \midrule
AFT         & 0.6028   & 62.83     & 0.5657 \\
AFT+Shift   & 4.9085   & 63.23     & 0.5604 \\
BiWKV+Shift & 6.8030   & 63.24     & 0.5551 \\ \bottomrule
\end{tabular}
\caption{Ablation study on the effects of different attention mechanisms in Bi-RWKV blocks. The $\Delta$FLOPs only counts the operations of attention layer or shift layer. Test R-D loss is used to indicate performance gain. 
}
\label{tab:block-component}
\end{table}

\section{Conclusion}

In this paper, we introduced LALIC, a straightforward yet effective linear attention-based learned image compression architecture. LALIC incorporates a Bi-RWKV block with a BiWKV attention mechanism and depth-wise convolution, enabling it to capture both local and global context effectively, thereby reducing redundancy. Furthermore, we applied the Bi-RWKV block to enhance channel-wise entropy modeling. Experimental results demonstrate that LALIC achieves superior rate-distortion performance on commonly used datasets, outperforming VTM-9.1 by -14.84\% in BD-rate on Kodak, Tecnick and CLIC Professional validation datasets. Compared to Convolution, Swin Transformer, and Mamba-based methods, LALIC still maintains efficiency in both decoding speed and parameter usage.

\section{Acknowledgment}

This work was partly supported by the Fundamental Research Funds for the Central Universities, STCSM under Grant 22DZ2229005, 111 project BP0719010. This work was also supported by Ant Group Research Fund.
{
    \small
    \bibliographystyle{ieeenat_fullname}

}

\clearpage
\setcounter{page}{1}
\maketitlesupplementary
\appendix

\section{Performance Details}

This section provides additional details regarding the results presented in Table~\ref{tab:performance}.

The rate-distortion (R-D) results can vary across different VTM anchors due to different evaluation process. To provide a generally accepted baseline on Kodak, we adopt the R-D results from the CompressAI repository~\cite{Begaint.2020.CompressAI}, which are collected from VTM-9.1. For other datasets, we use the following script to evaluate images in YUV space using VTM-9.1, where QPs range from 22, 27, 32, 37, 42, 47.

\begin{verbatim}
VTM/bin/EncoderAppStatic -i [input.yuv]
  -c VTM/cfg/encoder_intra_vtm.cfg
  -o [output.yuv] -b [output.bin]
  -wdt [width] -hgt [height] -q [QP] 
  --InputBitDepth=8 -fr 1 -f 1
  --InputChromaFormat=444
\end{verbatim}

Regarding runtime, FAT is reported to have a decoding time of 242 ms; however, our tests indicate a significantly longer decoding time of 426 seconds. This discrepancy remains an unresolved issue documented in its GitHub repository. Based on our analysis, decoding a single slice with FAT's T-CA entropy model involves computing masked channel attention across 12 layers, whereas TCM requires only two layers for decoding each slice. Incorporating techniques such as KV caching could potentially reduce the FLOPs required for each slice during decoding. Furthermore, the authors have acknowledged that the entropy coder in FAT requires additional optimization to improve decoding efficiency.

For complexity measurements, we use the \textit{thop} Python package to calculate parameters and FLOPs, ensuring consistency with the methodology employed for TCM~\cite{Liu.2023.TCM}. However, \textit{thop} has known limitations: it cannot account for FLOPs arising from non-layer-specific operations such as mathematical functions, matrix multiplications (e.g., in attention mechanisms), or CUDA-specific implementations. While the majority of FLOPs originate from Torch integrated layers, the values reported in Table~\ref{tab:performance} provide a reasonable and fair reference for comparison.

\section{Additional Experiment Results}

This section presents additional experimental results comparing our method with recent learned image compression (LIC) approaches. We present the BD-rate (MS-SSIM) results in Table~\ref{tab:ssim-bdrate-clic}, with VTM-9.1 as anchor. In fact, only a few recent works have publicly available MS-SSIM optimized models or corresponding curves on Tecnick/CLIC, resulting in some missing results.

\begin{table}
\centering
\begin{tabular}{lrrr}
\hline
Method                          & Kodak             & CLIC              & Tecnick           \\ \hline
VTM-9.1                         & 0.00\%            & 0.00\%            & 0.00\%            \\
ELIC        & -7.57\%           & -                 & -                 \\
TCM-large   & -49.76\%          & -                 & -                 \\
MLIC++ & -52.99\% & -47.43\% & -53.14\% \\
FAT          & -51.64\%          & -                 & -                 \\
LALIC (Ours)                    & -51.23\%          & -46.97\%          & -49.47\%          \\ \hline
\end{tabular}
\caption{BD-rate (MS-SSIM) performance relative to VTM-9.1 across different datasets. "-" indicates an unavailable result.}
\label{tab:ssim-bdrate-clic}
\end{table}

\section{Linear Attention Mechanisms}

Except the vanilla Attention which has a quadratic complexity, common modules have a linear complexity, including convolution, window-based attention. The recent linear attention methods, RWKV and Mamba are widely recognized for their efficiency in handling large-scale sequences to get a global reception filed, and also maintains the linear complexity with respect to the input size.

To provide a clearer comparison of these methods, Table~\ref{tab:attention-complexity} summarizes the theoretical time complexity of various attention mechanisms and the typical values of their number of operations (\#OPs).

\begin{table}[h!]
\centering
\begin{tabular}{@{}lrr@{}}
\toprule
Methods                    & Time Complexity          & \#OPs \\ \midrule
AFT~\cite{Zhai.2021.AFT}                 & \( 7LD \)          & \( 7LD \)   \\
AFT+Shift                  & \( 7LD + 50LD \)    & \( 57LD \)  \\
BiWKV+Shift                & \( 29LD + 50LD \)   & \( 79LD \)  \\
Window Attention~\cite{Liu.2021.SwinT}   & \( 2w^2LD \, (w=8) \) & \( 128LD \) \\
Selective Scan~\cite{Gu2023MambaLS}      & \( 9NLD \, (N=16) \)  & \( 144LD \) \\
Selective Scan 2D~\cite{Liu2024VMambaVS} & \( 4 \times 9NLD \, (N=16) \) & \( 576LD \) \\ \bottomrule
\end{tabular}
\caption{Theoretical time complexity of various attention mechanisms in terms of number of operations (\#OPs).}
\label{tab:attention-complexity}
\end{table}

In all cases, the computational cost is directly proportional to \( L \cdot D \), where \( L \) represents the sequence length, and \( D \) denotes the latent dimension. The theoretical FLOPs for various mechanisms are outlined below:

\begin{itemize}    
    \item \textbf{AFT+Shift}: The complexity of the AFT (named AFT-simple in ~\cite{Zhai.2021.AFT}) is estimated as \( 7LD \) by the \textit{torch-operation-counter} package. Adding the 5x5 depth-wise convolution shift operation (\( 25LD \times 2 \) for both spatial and channel mix modules) increases the total complexity to \( 57LD \).
    
    \item \textbf{BiWKV+Shift}: The BiWKV~\cite{Duan.2024.Vision-RWKV} mechanism, computed as \( 29LD \) according to the Vision-RWKV GitHub repository, combined with the shift operation results in \( 79LD \).
    
    \item \textbf{Window Attention}: The window attention~\cite{Liu.2021.SwinT} mechanism has a complexity of \( 2w^2LD \), where \( w \) is the window size, typically set to 8, resulting in \( 128LD \).
    
    \item \textbf{Selective Scan}: In Mamba~\cite{Gu2023MambaLS}, the selective scan mechanism has a complexity of \( 9NLD \), where \( N \) is the state dimension, typically set to 16, leading to \( 144LD \). In SS2D~\cite{Liu2024VMambaVS}, the selective scan is performed four times, resulting in a total complexity of \( 4 \times 9NLD = 576LD \).
\end{itemize}

As shown in Table~\ref{tab:attention-complexity}, BiWKV attention demonstrates significant computational efficiency compared to these other mechanisms, making it a compelling choice for balancing performance and complexity.

\section{Linear Complexity on Scaling}

\begin{figure}[h!]
    \centering
    \begin{subfigure}[b]{0.45\textwidth}
        \includegraphics[width=\textwidth]{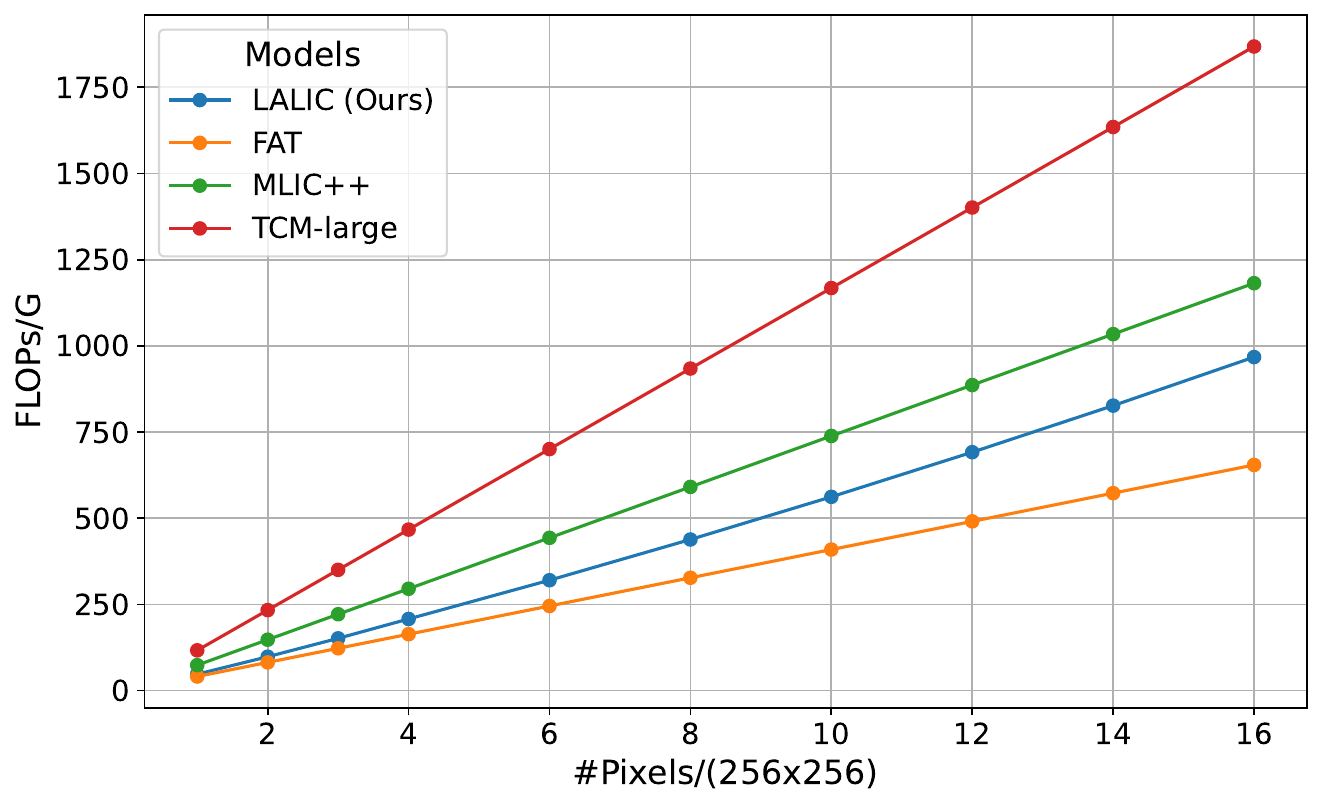}
        \caption{FLOPs vs. Resolution}
        \label{fig:sub1}
    \end{subfigure}
    \begin{subfigure}[b]{0.45\textwidth}
        \includegraphics[width=\textwidth]{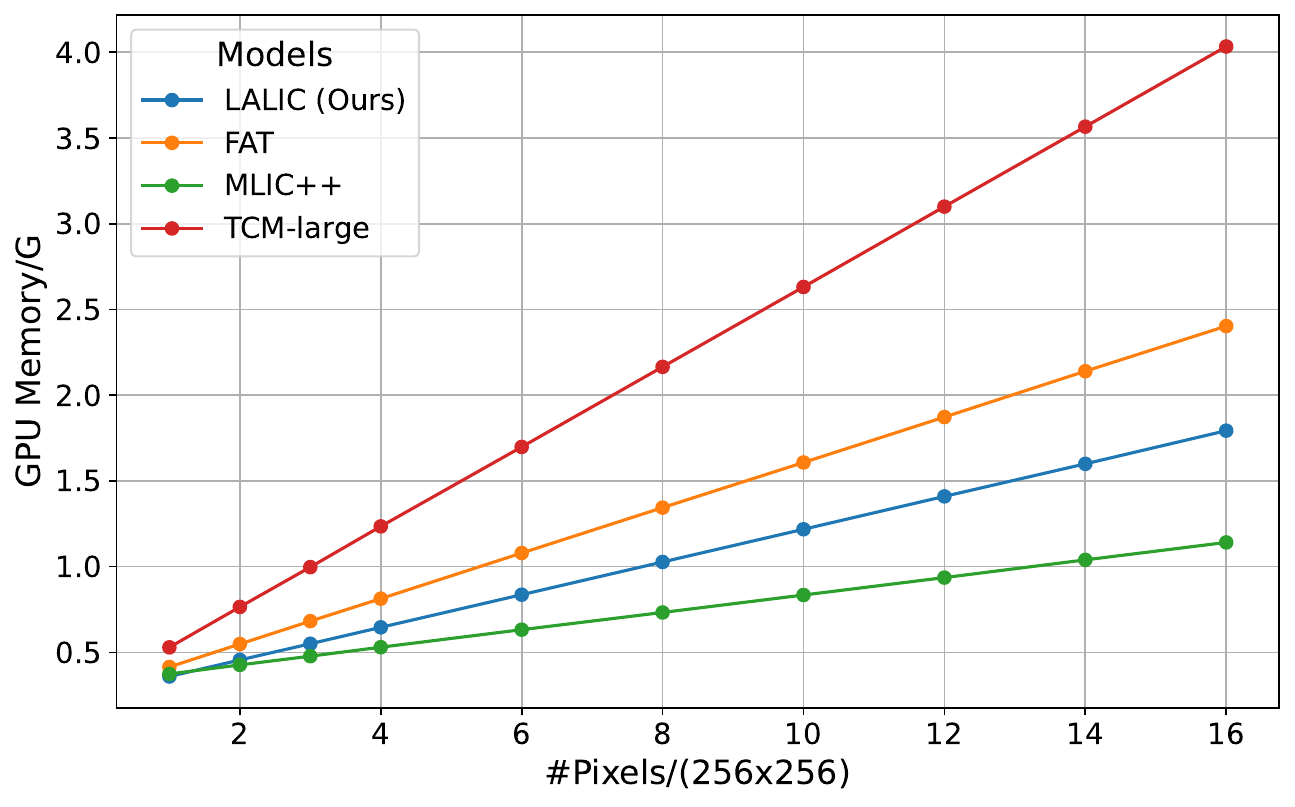}
        \caption{GPU Memory Usage vs. Resolution}
        \label{fig:sub2}
    \end{subfigure}
  \caption{Linear scaling trends of FLOPs (a) and GPU memory usage (b) for different LIC methods as a function of image resolution. LALIC achieves competitive performance with medium-low computational and memory demands.}
  \label{fig:trend-linear}
\end{figure}

Practical learned image compression (LIC) methods exhibit linear complexity with respect to the number of pixels, as shown in Figure~\ref{fig:trend-linear}. Unlike previous demonstrations~\cite{Jiang.2023.MLICpp} that used a quadratic x-axis and presented a quadratic trend for all methods, this figure employs a linear x-axis for clarity, providing a more intuitive understanding for readers. The maximum resolution tested is 1024$\times$1024.

Among recent LIC methods, our proposed LALIC demonstrates medium-low FLOPs and forward GPU memory usage, striking a balance between computational efficiency and memory requirements.

\section{Entropy Model Architecture}

\begin{figure}[h!]
  \centering
  \includegraphics[width=\linewidth]{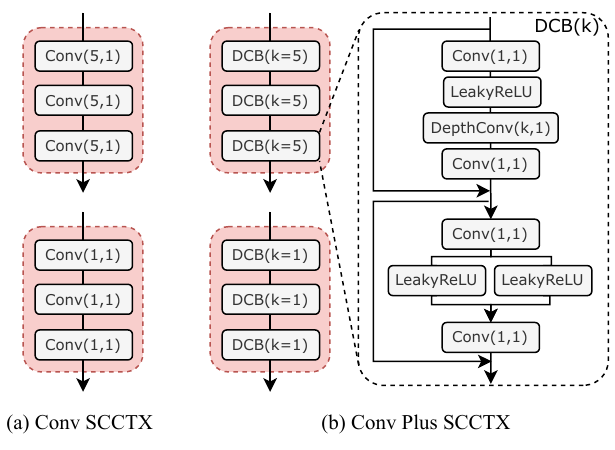}
  \caption{Network architecture of Conv SCCTX and Conv Plus SCCTX configurations. The upper module represents channel context extraction, while the lower module corresponds to parameter aggregation.}
  \label{fig:Conv-SCCTX}
\end{figure}

For entropy models, we adopt the Conv SCCTX model~\cite{He.2022.ELIC} and an enhanced Conv Plus SCCTX configuration as reference baselines. The detailed network architectures of these models are illustrated in Figure~\ref{fig:Conv-SCCTX}.

The Conv SCCTX model consists of three 5$\times$5 convolutional (Conv) layers designed to extract channel context, followed by three 1$\times$1 Conv layers for entropy parameter estimation. The Conv Plus SCCTX configuration extends this architecture by incorporating Depth Conv Block (DCB) from the DCVC\cite{Sheng.2023.TCM} learned video coding series, where the hyperparameter \( k \) denotes the kernel size of the depthwise convolution. To further enhance the modeling capacity, we increase the channel dimensions in the depthwise convolution layers, thereby raising the number of context parameters.

\begin{figure*}[h]
\centering
\subfloat[Original]{\includegraphics[width=2.15in]{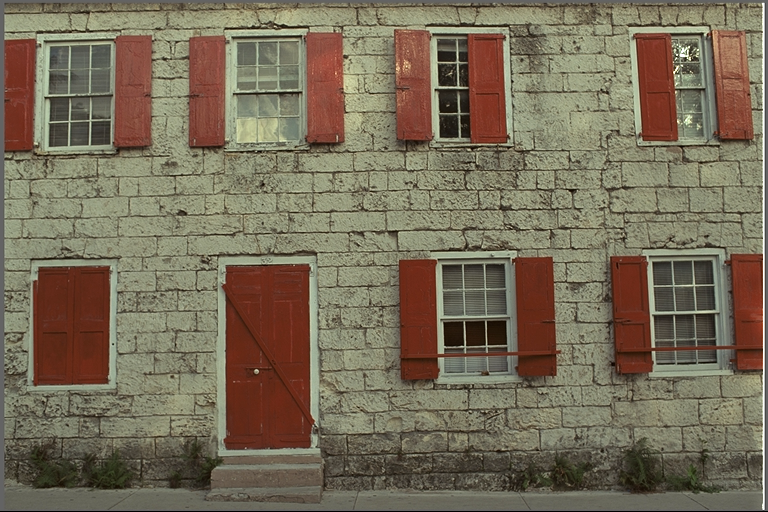}%
\label{k01-origin}}
\hfil
\subfloat[TCM-large 0.767 bpp / 32.07 dB]{\includegraphics[width=2.15in]{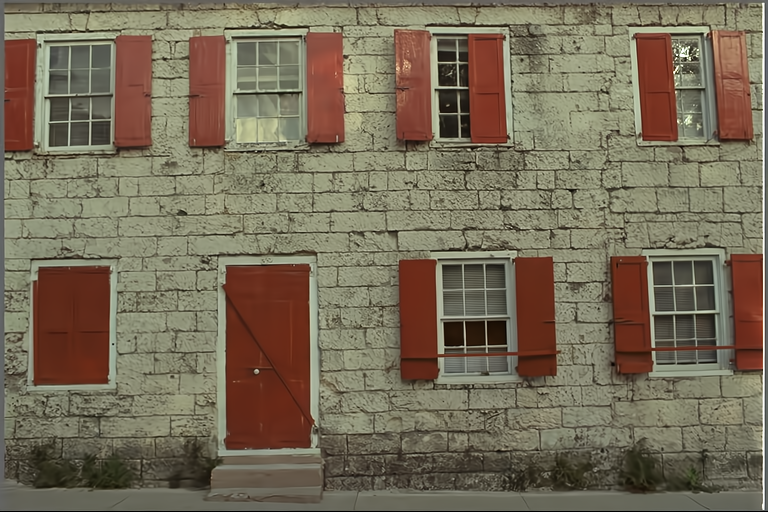}%
\label{k01-tcm}}
\hfil
\subfloat[LALIC (Ours) 0.759 bpp / 32.12 dB]{\includegraphics[width=2.15in]{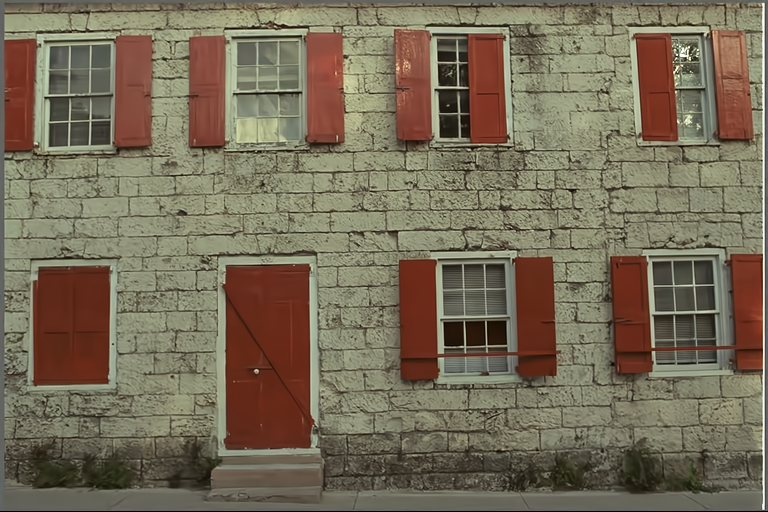}%
\label{k01-our}}
\hfil
\subfloat[Original crop]{\includegraphics[width=1.5in]{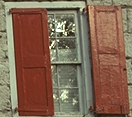}%
\label{k01-c_o}}
\hfil
\subfloat[TCM-large crop]{\includegraphics[width=1.5in]{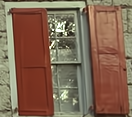}%
\label{k01-c_tcm}}
\hfil
\subfloat[LALIC (Ours) crop]{\includegraphics[width=1.5in]{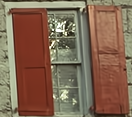}%
\label{k01-c_our}}
\captionsetup{font=small}
\caption{Subjective quality comparison on the $kodim01$ image from Kodak.}
\label{subjective01}
\vspace{-5pt}
\end{figure*}

\begin{figure*}[h]
\centering
\subfloat[Original]{\includegraphics[width=2.15in]{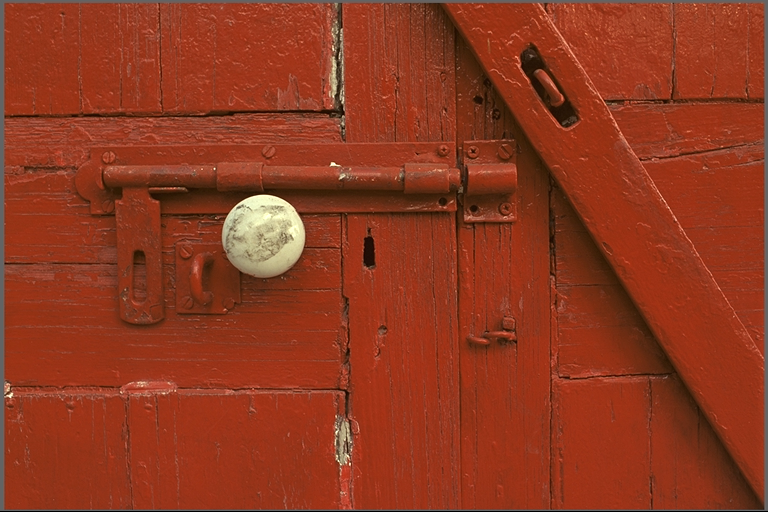}%
\label{k02-origin}}
\hfil
\subfloat[TCM-large 0.245 bpp / 34.20 dB]{\includegraphics[width=2.15in]{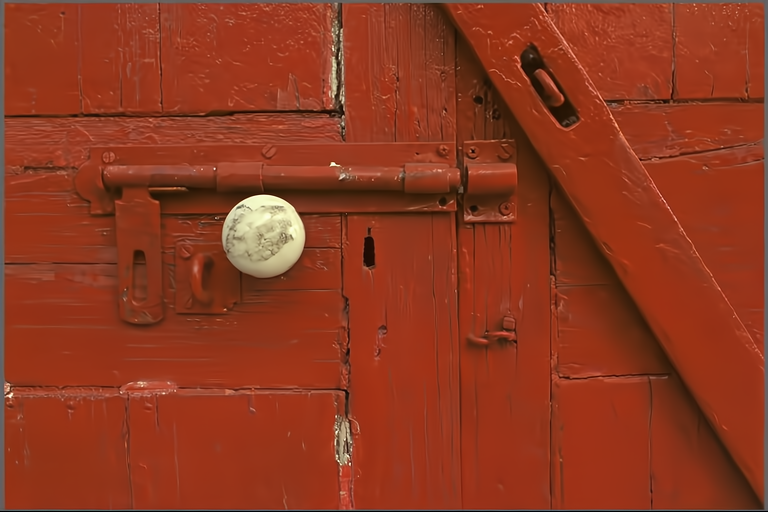}%
\label{k02-tcm}}
\hfil
\subfloat[LALIC (Ours) 0.243 bpp / 34.30 dB]{\includegraphics[width=2.15in]{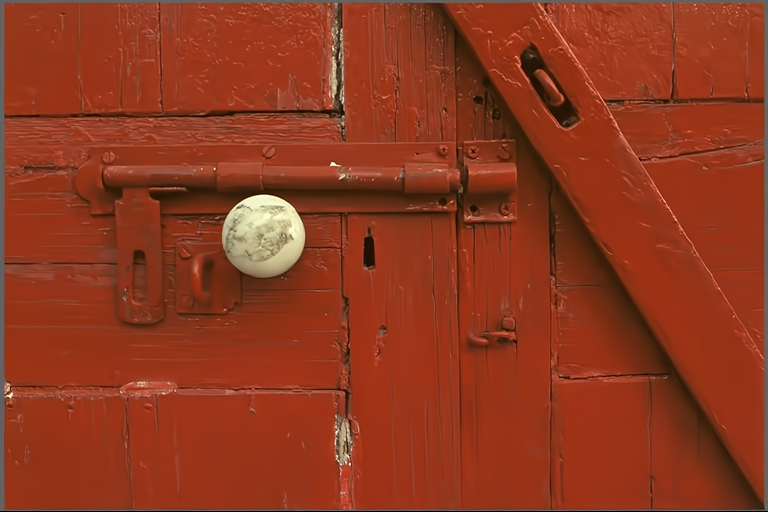}%
\label{k02-our}}
\hfil
\subfloat[Original crop]{\includegraphics[width=1.2in]{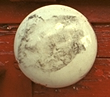}%
\label{k02-c_o}}
\hfil
\subfloat[TCM-large crop]{\includegraphics[width=1.2in]{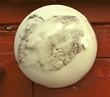}%
\label{k02-c_tcm}}
\hfil
\subfloat[LALIC (Ours) crop]{\includegraphics[width=1.2in]{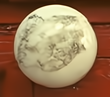}%
\label{k02-c_our}}
\captionsetup{font=small}
\caption{Subjective quality comparison on the $kodim02$ image from Kodak.}
\label{subjective02}
\vspace{-5pt}
\end{figure*}

\section{Subjective Results}

We conducted a subjective comparison of reconstructed images generated by our LALIC model and our trained TCM-large model on the Kodak dataset. The results are shown in Figure~\ref{subjective01} and Figure~\ref{subjective02}. By focusing on specific image regions, we observe that our proposed method preserves finer details compared to TCM-large. For instance, LALIC retains sharper textures in the wooden board on the right side of Figure~\ref{subjective01} and captures the intricate structure of the door handle in Figure~\ref{subjective02}.

In addition to qualitative improvements, our method achieves higher PSNR values while maintaining a lower bitrate, highlighting its superior rate-distortion performance over TCM-large.

\end{document}